\documentclass[10pt,twocolumn,letterpaper]{article}

\usepackage{iccv}
\usepackage{times}
\usepackage{epsfig}
\usepackage{graphicx}
\usepackage{amsmath}
\usepackage{amssymb}
\usepackage{times}
\usepackage{epsfig}
\usepackage{graphicx}
\usepackage{amsmath}
\usepackage{amssymb}
\usepackage{mathrsfs}
\usepackage{multirow}
\usepackage{booktabs}

\usepackage{algorithm}
\usepackage{algorithmic}
\usepackage{bm}
\usepackage{color}
\usepackage{amssymb}
\usepackage{latexsym}
\usepackage{amsmath}
\usepackage{mathrsfs}
\usepackage{graphicx}
\usepackage{float}
\usepackage{subfigure}
\usepackage{caption}

\usepackage[pagebackref=true,breaklinks=true,letterpaper=true,colorlinks,bookmarks=false]{hyperref}

\iccvfinalcopy 

\def\iccvPaperID{4} 

\ificcvfinal\fi

\begin{document}

\title{A Cascaded Zoom-In Network for Patterned Fabric Defect Detection}

\author{Zhiwei Zhang\footnotemark[1]\\
{\tt\small bitzzw@gmail.com}
}

\maketitle
\ificcvfinal\thispagestyle{empty}\fi

\begin{abstract}
   Nowadays, Deep Convolutional Neural Networks (DCNNs) are widely used in fabric defect detection, which come with the cost of expensive training and complex model parameters. With the observation that most fabrics are defect free in practice, a two-step Cascaded Zoom-In Network (CZI-Net) is proposed for patterned fabric defect detection. In the CZI-Net, the Aggregated HOG (A-HOG) and SIFT features are used to instead of simple convolution filters for feature extraction. Moreover, in order to extract more distinctive features, the feature representation layer and full connection layer are included in the CZI-Net. In practice, Most defect-free fabrics only involve in the first step of our method and avoid a costive computation in the second step, which makes very fast fabric detection. More importantly, we propose the Locality-constrained Reconstruction Error (LCRE) in the first step and Restrictive Locality-constrained Coding (RLC), Bag-of-Indexes (BoI) methods in the second step. We also analyse the connections between different coding methods and conclude that the index of visual words plays an essential role in the coding methods. In conclusion, experiments based on real-world datasets are implemented and demonstrate that our proposed method is not only computationally simple but also with high detection accuracy.
\end{abstract}

\renewcommand{\thefootnote}{\fnsymbol{footnote}}
	\footnotetext[1]{Technical report without peer review. This work was completed independently by Zhiwei Zhang.}

\section{Introduction}
Defect detection is critical to the quality control of fabrics. The efficiency of traditional manual inspection is low and the missed rate is high because of eye fatigue. Therefore, fabric defect detection based on computer vision is more popular in practice~\cite{ICSP2018}. In the literature, fabric defect detection methods were categorized into seven groups: statistical, spectral, model-based, learning, structural, hybrid and motif-based \cite{NganPang2011}. Spectral approaches are the most popular method for fabric defect detection, they address defect detection in the frequency domain by using tools such as Fourier transform (FT) \cite{Chan2000} Wavelet transform (WT) \cite{Kim1999}, Gabor transform (GT) \cite{Kumar2002}. However, it is not suitable to use spectral approaches for fabric defect detection containing random texture.

As deep learning attracts significant attention, learning approaches based on deep convolutional neural networks (DCNNs), performing automated feature learning instead of hand-designing suitable features, are widely used and demonstrate excellent performance on fabric defect detection. For example, \cite{LiZhaoPan2017} classified the defect-free and defective fabrics by using Fisher criterion-based stacked denoising autoencoders (FCSDA). Experimental results showed that the FCSDA method could obtain 99.47\% accuracy on complex jacquard warp-knitted fabric. \cite{Racki2018} designed a global pyramid-fashion v1173 ($11 \times 11$, $7 \times 7$, $3 \times 3$) network architecture by decreasing the filter size in each subsequent convolution block for textured-surface defect segmentation and detection, which achieved nearly 100\% detection accuracy. \cite{Racki2018} also compared the designed v1173 network architecture with v333 ($3 \times 3$, $3 \times 3$, $3 \times 3$) network architecture and concluded that the designed v1173 network architecture accounted for local context in a pyramid fashion performed better. Although v1173 network architecture is more invariant to image transformation, it has more than one million parameters to be optimized. Additionally, DCNNs-based feature extraction is a data-driven technique that needs lots of images for training the complex model parameters, which limits its application on fabric defect detection. Especially, DCNNs-based feature extraction is expensive for the patterned fabric with simple patterns. Therefore, we look forward to finding feature extractors that are more capable of coping with image transformations.

Fortunately, Scale-Invariant Feature Transform (SIFT) \cite{Lowe2004} and Histograms of Oriented Gradients (HOG) \cite{DalalTriggs2005} can extract more stable features than simple convolution filters. In our method, the Aggregated HOG (A-HOG) feature is proposed for feature extraction, which is a high dimension vector aggregated from extracted HOG features in the image sub-region. The extraction of A-HOG feature is fast, but its disadvantage is that it can't determine the location of defect in the image sub-region. SIFT feature that invariant to scale and rotation can detect tiny defects, but high computational complexity limits its application. Additionally, With the observation that most fabrics are defect free in practice, a two-step Cascaded Zoom-In Network (CZI-Net) is proposed to explore the possibility of combing the merits of both the A-HOG and the SIFT features. In the CZI-Net, A-HOG feature-based feedforward network (AHOG-Net) and SIFT feature-based feedforward network (SIFT-Net) are included for fabric defect detection. The flowchart of CZI-Net is illustrated in Fig. 1. In the AHOG-Net, image sub-regions with several texture primitives are detected to determine whether there is defective. If the image reconstruction error is less than the threshold, there is no need to enter the SIFT-Net. Otherwise, the SIFT-Net is adopted where the texture primitives are evaluated one by one. For most fabrics, only the AHOG-Net is used and the computation cost is reduced. For those image sub-regions with defects, the detection result is consistent ensured by the adoption of SIFT-Net.

\begin{figure}[!t]
\centering
\includegraphics[scale=.58]{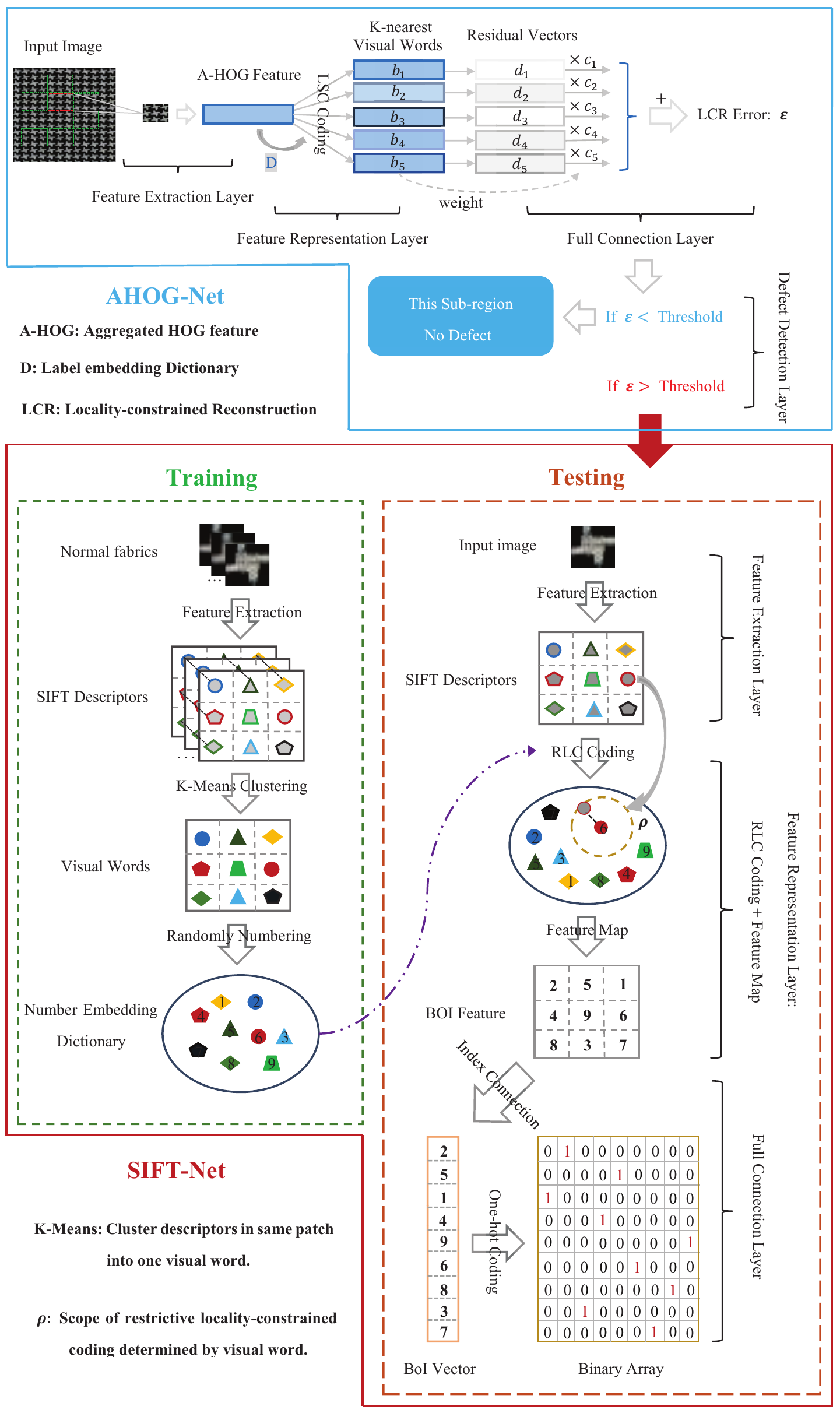}
\caption{The flowchart of two-step Cascaded Zoom-In Network.}
\end{figure}

As shown in Fig. 1, both the AHOG-Net and the SIFT-Net include feature representation layer and full connection layer for enhancing the image representation capability. (i) In the AHOG-Net, Localized Soft-assignment Coding (LSC) \cite{LiuWangLiu2011} is used to obtain the \textit{k}-nearest visual words of local descriptors and their weights. For fabric defect detection, the Locality-constrained Reconstruction Error (LCRE) is proposed, which is the linear combination of weighted residual vectors. The residual vectors are the difference between the local descriptors and their \textit{k}-nearest visual words. Finally, compare the LCRE value with the threshold to determine whether there is defective in the sub-regions of fabric. If there are defects in sub-regions, the SIFT-Net is adopted for further detection. (ii) In the SIFT-Net, the proposed Restrictive Locality-constrained Coding (RLC) and Bag-of-Indexes (BoI) methods together form the feature representation layer. RLC is the extension of Locality-constrained Linear Coding (LLC) \cite{WangYangYuHuangGong2010}. LLC utilizes the locality constrains to project each descriptor into its local-coordinate system, and descriptor is represented by its k-nearest visual words in the dictionary. However, LLC doesn’t consider the degree of similarity between the descriptor and its k-nearest visual words. Therefore, the proposed RLC limits the scope of locally encoding, which means that only the visual words with in this local scope are valid for encoding. Based on the proposed the RLC, the nearest visual word of each descriptor will be signed for feature map. In the feature map step, the numbered visual word will be mapped to obtain its index in the dictionary. In this step, we also propose the BoI feature based on the Bag-of-Features (BoF) \cite{CsurkaDanceFan2004}. The BoF method represents an image as a histogram of its local features, which disregards the information about the spatial layout of features. In order to overcome the disadvantage of BoF, the Spatial pyramid matching (SPM) method \cite{LazebnikSchmidPonce2006} partitions the image into increasingly finer spatial sub-regions and computes histograms of local descriptors from each sub-region. We can observe that if each patch of SIFT feature is considered as a sub-region, the BoF feature will be transformed into BoI feature. BoI feature means that the order of visual words is used to overcome the disadvantage of orderless in BoF. In section 4, our experiments show that BoI feature performs better than the BoF feature. And then in the full connection layer, the BoI features are fully connected to form the BoI vector, and one-hot encoding is used to extend the discrete index features to the European space. A certain value of the discrete index corresponds to a certain point in the binary array. Finally, in the defect detection layer, the Hamming distance between binary array and the scope of locally encoding are compared with thresholds for defect detection. In addition, it should be noted that the label embedding dictionary  is used in the AHOG-Net. And in the training state of SIFT-Net, the SIFT features in the same patch are clustered by K-Means to obtain the visual words, and the visual words are randomly numbered to form the number embedding dictionary.

Our contributions are listed as follows. (i) In this paper, we propose a two-step Cascaded Zoom-In Network (CZI-Net) for patterned fabrics defect detection, which includes AHOG-Net and SIFT-Net. (ii) The Locality-constrained Reconstruction Error (LCRE) is proposed in the AHOG-Net. (ii) The Restrictive Locality-constrained Coding (RLC) and Bag-of-Indexes (BoI) methods are proposed in the SIFT-Net. Experiments demonstrate that the BoI method performs better than the BoF method, and the index of visual word plays an important role in the coding methods. (iii) Experiments are implemented based on real-world dataset to evaluate the proposed CZI-Net and achieve state-of-the-art performance.

The remainder of this paper is organized as follows. In Section 2, we provide some background knowledge on DCNNs, coding methods and BoF method. Section 3 introduces the proposed CZI-Net network. In section 4, experiments are implemented. Finally, we conclude our work in Section 5.

\section{Preliminary}
In this section, we provide some background knowledge of our method. In subsection 2.1, we analyse the disadvantage of redundant convolution filters in DCNNs for feature extraction, and use HOG and SIFT features to instead of it for feature extraction. In subsection 2.2, coding methods are introduced. In subsection 2.3, we introduce the BoF method and analysis its disadvantages.

\subsection{Deep Convolutional Neural Networks}
Deep Convolutional Neural Networks (DCNNs), based on convolution filters, have attracted significant attentions in computer vision due to its amazing powerful feature representation capabilities from raw image pixels. However, complex network architecture limits its application in practical fabric defect detection. As we know, traditional DCNNs include convolutional layers, pooling layers, and fully connected layers, such as LeNet \cite{LeCunBottouBengioHaffner1998}, AlexNet \cite{KrizhevskySutskeverHinton2012}, VGGNet \cite{SimonyanZisserman2014}. A typical flowchart of DCNNs is illustrated in Fig. 2. In this figure, there are three layers: input layer, hidden layer and output layer. The hidden layer can be considered as feature extraction layer consisting of two convolutional layers (C1 Layer, C2 Layer), two pooling layers (P1 Layer, P2 Layer), and one fully connected layer (FC Layer). \cite{LuanChenZhangHanLiu2018} had concluded that the learned filters are redundant in the hidden layers of the AlexNet. Here, it is valuable to note that redundant convolution filters are used to extract more discriminative features and lead to expensive training and complex model parameters. For example, max-pooling is utilized in a single scale of images, resulting in that more different size convolution filters are used to extract scale-invariant features. In this following, in order to enhance the model capacity to image transformations, HOG \cite{DalalTriggs2005} and SIFT features \cite{Lowe2004} can be used to instead of the simple convolution filters in DCNNs for feature extraction.

\begin{figure}[!t]
\centering
\includegraphics[scale=.6]{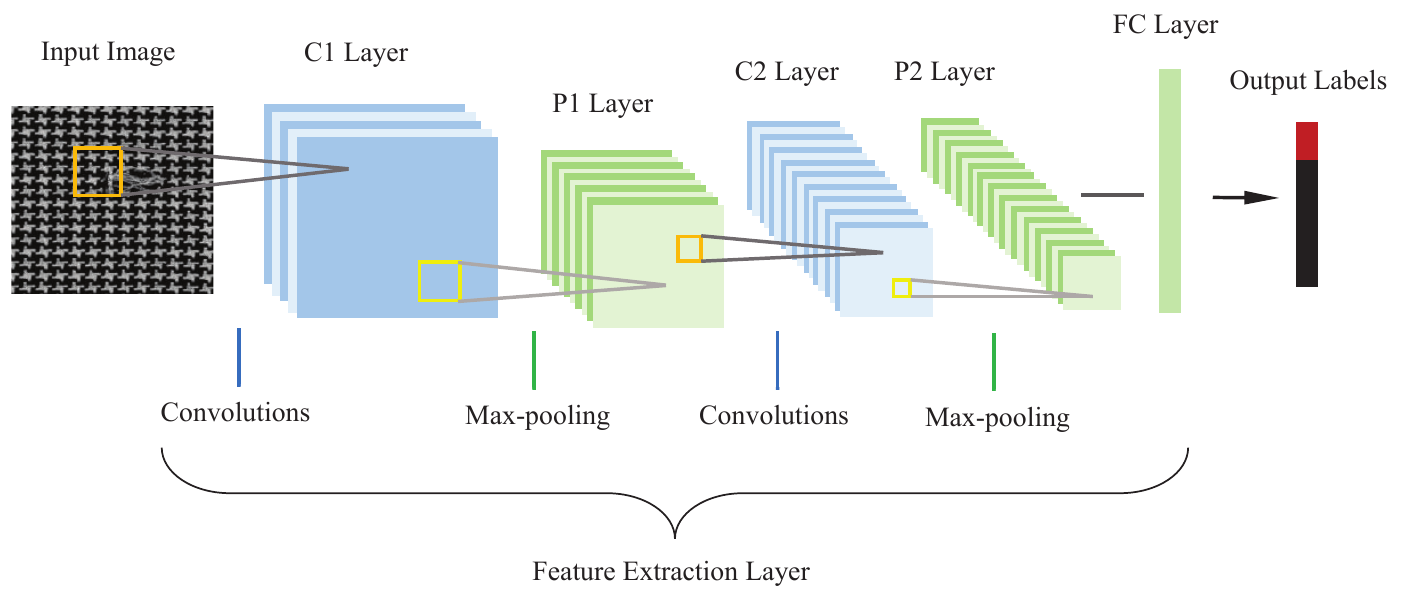}
\caption{The flowchart of traditional deep convolutional neural networks for fabric defect detection.}
\end{figure}

\subsection{Coding methods}
Let $\textit{\textbf{X}} = [{\textit{\textbf{x}}_1}, {\textit{\textbf{x}}_2}, ... , {\textit{\textbf{x}}_N}] \in \mathbb{R}^{\textit{D} \times \textit{N}}$ be a set of D-dimensional local features extracted from an image, $\textit{\textbf{B}} = [{\textit{\textbf{b}}_1}, {\textit{\textbf{b}}_2}, ... , {\textit{\textbf{b}}_M}] \in \mathbb{R}^{\textit{D} \times \textit{M}}$ be a M-entry codebook generated from K-Means clustering algorithm.
Local features $\textit{\textbf{X}}$ is converted to $\textit{N}$ coding vectors $\textit{\textbf{C}} = [{\textit{\textbf{c}}_1}, {\textit{\textbf{c}}_2}, ... , {\textit{\textbf{c}}_N}] \in \mathbb{R}^{\textit{M} \times \textit{N}}$ by encoding. The following subsection reviews the existing coding methods.

\subsubsection{Hard-assignment Coding}
Local feature $\textbf{\textit{x}}_i$ is signed to its nearest visual word. There is one and only one nonzero coding coefficient. When Euclidean distance is used,
\begin{equation}
{\textit{c}}_{ij} =
\begin{cases}
1& if j = \mathop{\arg\min\limits_j} \|\textbf{\textit{x}}_i - \textbf{\textit{b}}_j\|^2  \\
0& \text{otherwise}
\end{cases}
\end{equation}

\subsubsection{Soft-assignment Coding}
Soft-assignment coding \cite{VanGemert2008} represents a local feature by multiple visual words, and their degree of membership is calculated by the kernel function (e.g., the Gaussian function). The \textit{j}th coding coefficient is:
\begin{equation}
{\textit{c}}_{ij} = \frac{\exp(-\beta\|\textbf{\textit{x}}_i - \textbf{\textit{b}}_j\|^2)}{\sum_{j=1}^{M}\exp(-\beta\|\textbf{\textit{x}}_i - \textbf{\textit{b}}_j\|^2)}
\end{equation}
where $\beta$ is the smoothing factor controlling the softness of the assignment. Note that all the $\textit{M}$ visual words are used in the computing ${\textit{c}}_{ij}$.

\subsubsection{Localized Soft-assignment Coding}
\cite{LiuWangLiu2011} concluded that the performance of soft-assignment coding \cite{VanGemert2008} is sensitive with $\beta$ ($\exp(-\beta\| \textbf{\textit{x}}_i - \textbf{\textit{b}}_j\|^2$). Because $\beta$ may not be reliably estimated for every Gaussian function. To solve this problem, \cite{LiuWangLiu2011} proposed the Localized Soft-assignment Coding (LSC) that only considers the \textit{k}-nearest visual words of local descriptors and conceptually sets its distances to the remaining words as infinity. This strategy produces an ``early cut-off" effect, which removes the adverse impact of unreliable longer distance even if a small $\beta$ is used. Locality soft-assignment coding can be written as follows:
\begin{equation}
{\textit{c}}_{ij} = \frac{K_\beta(\textbf{\textit{x}}_i, \textbf{\textit{b}}_j)}{\sum_{j=1}^{k}K_\beta(\textbf{\textit{x}}_i, \textbf{\textit{b}}_j)}
\end{equation}
\begin{equation}
K_\beta(\textbf{\textit{x}}_i, \textbf{\textit{b}}_j) =
\begin{cases}
\exp(-\beta\|\textbf{\textit{x}}_i - \textbf{\textit{b}}_j\|^2)& \text{if $\textbf{\textit{b}}_j \in \mathcal{N}_k(\textbf{\textit{x}}_i)$}\\
\ \ \ \ \ \ \  \ \ \ \ \ \ 0& \text{otherwise}
\end{cases}
\end{equation}
where $\mathcal{N}_k(\textbf{\textit{x}}_i)$ denotes the \textit{k}-nearest visual words of $\textbf{\textit{x}}_i$.

\subsubsection{Locality-constrained Linear Coding}
The Locality-constrained Linear Coding (LLC) \cite{WangYangYuHuangGong2010} encodes each local descriptor ${\textit{\textbf{x}}}_{\textit{i}}$ by solving the following problem:
\begin{equation}
\begin{split}
&{}\mathop{\min}_{C} \sum\limits_{i=1}^{N} \ \|\ {\textbf{\textit{x}}_i - \textbf{\textit{B}}\textbf{\textit{c}}_i\ \|}^2 + \lambda \|\ {\textbf{\textit{d}}_i \odot \textbf{\textit{c}}_i\ \|}^2 \\
&{} s.t. \ {\textbf{1}^\top}{\textbf{\textit{c}}_i} = 1, \forall{i} \\
\end{split}
\end{equation}
\begin{equation}
\textbf{\textit{d}}_i = \exp\bigg(\frac{\textit{dist}(\textbf{\textit{x}}_i, \textbf{\textit{B}})}{\sigma}\bigg)
\end{equation}
\begin{equation}
\begin{split}
&{}\tilde{\textbf{\textit{c}}}_i = \big{(}\textbf{\textit{C}}_i + \lambda \rm{diag(d)}\big{)} \setminus 1 \\
&{}\textbf{\textit{c}}_i = \tilde{\textbf{\textit{c}}}_i / {\textbf{1}^\top}{\textbf{\textit{c}}_i} \\
\end{split}
\end{equation}
where $\odot$ denotes the element-wise multiplication, $\textbf{\textit{d}}_i \in \mathbb{R}^\textit{M}$ is the locality adaptor that gives different freedom for each basis vector proportional to its similarity to the input descriptor $\textbf{\textit{x}}_i$. $\textit{dist}(\textbf{\textit{x}}_i, \textbf{\textit{B}})= [\textit{dist}(\textbf{\textit{x}}_i, \textbf{\textit{b}}_1), \textit{dist}(\textbf{\textit{x}}_i, \textbf{\textit{b}}_2), ... , \textit{dist}(\textbf{\textit{x}}_i, \textbf{\textit{b}}_M)]^{T}$, $\textit{dist}(\textbf{\textit{x}}_i, \textbf{\textit{b}}_j)$ is the Euclidean distance between $\textbf{\textit{x}}_i$ and $\textbf{\textit{b}}_j$, $\sigma$ is used for adjusting the weight decay speed for the locality adaptor. The constraint ${\textbf{1}^{T}}{\textbf{\textit{c}}_i} = 1$ follows the shift-invariant requirements of the LLC codes. $\textbf{\textit{C}}_i = (\textbf{\textit{B}}-\textbf{1}{\textbf{\textit{x}}_i}^\top){(\textbf{\textit{B}}-\textbf{1}{\textbf{\textit{x}}_i}^\top)}^\top$ denotes the data covariance matrix.

In practice, the approximated LLC is used for fast encoding. It use the \textit{K} $(\textit{K} < \textit{D} < \textit{M})$ nearest neighbors of $\textbf{\textit{x}}_i$ as the local bases $\textbf{\textit{B}}_i$, and solve a much smaller linear system to get the codes:
\begin{equation}
\begin{split}
&{}\mathop{\min}_{\tilde{C}} \sum\limits_{i=1}^{N} \ \|\ {\textbf{\textit{x}}_i - {\textbf{\textit{B}}_i}\tilde{\textbf{\textit{c}}}_i\ \|}^2 \\
&{}s.t. \ {\textbf{1}^\top}{\tilde{\textbf{\textit{c}}}_i} = 1, \forall{i} \\
\end{split}
\end{equation}

In the solution of (8), although \textit{K} is very small, the dictionary is very large, which causes the high computational complexity of the LLC. To deal with this disadvantage, as shown in training state of Fig. 1, we construct the extracted descriptors into three-dimensional tensors, and then use the K-Means algorithm to cluster the descriptors in the same patch to obtain the corresponding visual words. This step makes the number of visual words in the dictionary equal to the number of descriptors, which not only reduces the size of dictionary, but also the detection accuracy is no longer dependent on the size of dictionary. Additionally, as is shown on the left of Fig. 6, LLC approach utilizes the locality constrains to project each descriptor into its local-coordinate system, and descriptor is represented by its k-nearest visual words in the dictionary. However, LLC doesn't consider the degree of similarity between the descriptor and its k-nearest visual words.

\subsection{Bag-of-Features}
The Bag-of-Features (BOF) \cite{CsurkaDanceFan2004} method is extremely popular in computer vision. The method treats an image as a collection of unordered appearance descriptors extracted from local patches, quantizes them into discrete "visual words", and then computes a compact histogram representation for semantic image classification. However, The BoF method disregard the information about the spatial layout of features, which limits the descriptive power of the image representation.

The flowchart of the BOF method is illustrated in Fig. 3. BoF method also includes training state and testing state. In the training state, all the local descriptors extracted from normal fabrics are clustered by K-Means algorithm to obtain the dictionary. In the testing state, the local descriptors extracted from the tested images are signed to the nearest visual words by hard coding \cite{HuangTan2014}. Next, the frequencies of the signed visual words are recorded to form the BoF histogram. Finally, the histograms of images are used for classification, such as calculate the distance of histograms or put the histograms into support vector machine (SVM) \cite{LazebnikSchmidPonce2006}.

\begin{figure}[!t]
\centering
\includegraphics[scale=.6]{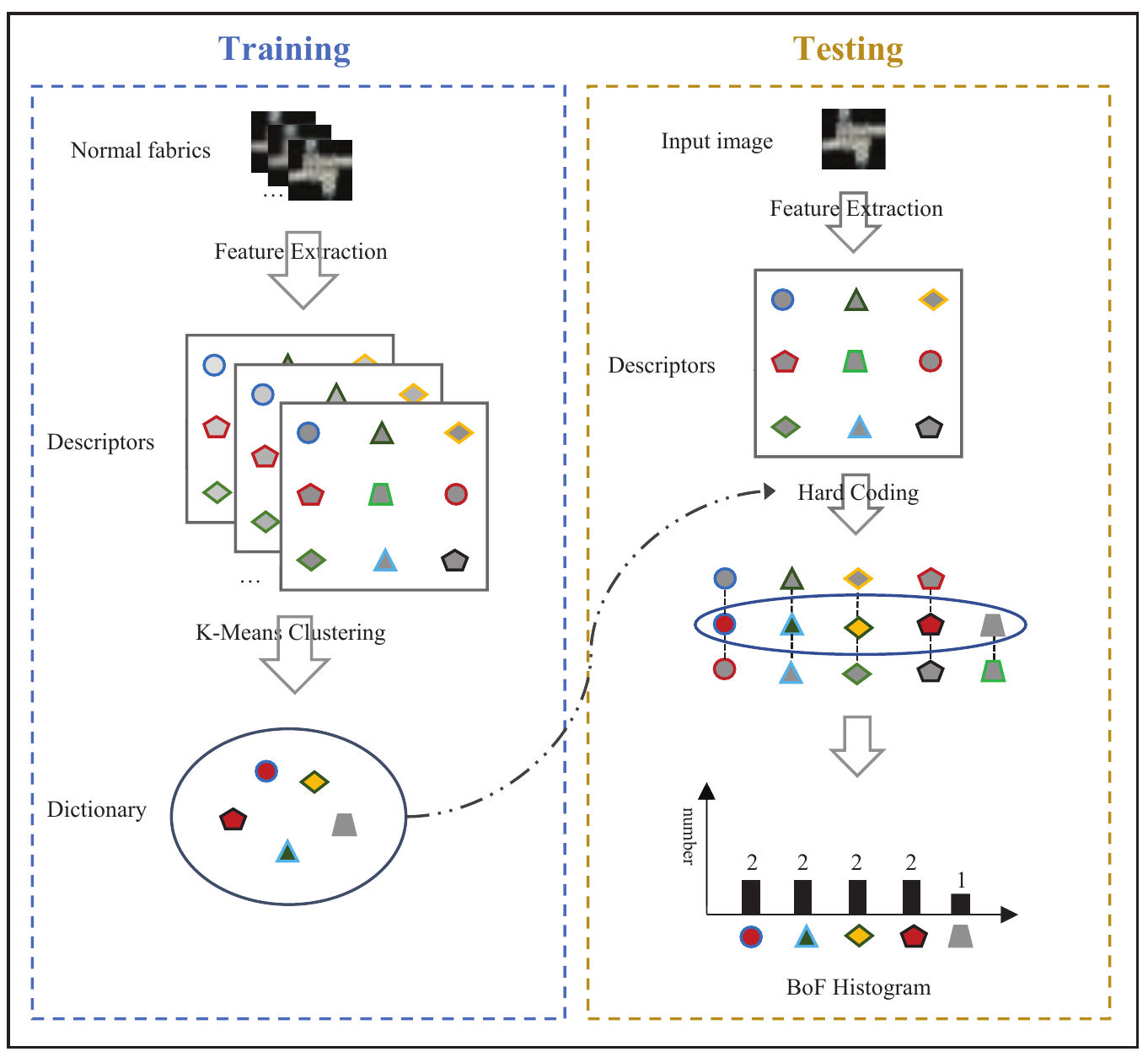}
\caption{The flowchart of the BoF method.}
\end{figure}

\section{The proposed method}
In this section, we proposed a Cascaded Zoom-In Network (CZI) for fabric defect detection. In subsection 3.1, the method of fabric image segmentation is introduced. Subsection 3.2 presents the proposed AHOG-Net and Locality-constrained Reconstruction Error (LCRE). Subsection 3.3 presents the proposed SIFT-Net and Restrictive Locality-constrained Coding (RLC), Bag-of-Indexes (BoI) methods.

\subsection{Fabric image segmentation}
As shown in Fig. 4, there is lots of noise on fabric image before smoothing. Therefore, in order to reduce the influence of noise, the Gaussian filter is used to smooth the fabric image before segmentation. After smoothing, the texture primitives can be more easily segmented. In our CZI-Net method, the fabric image segmentation is divided into two step corresponding to the AHOG-Net and SIFT-Net defect detection. As shown in Fig. 5, a fabric image is segmented into 12 sub-regions for detection. A sub-region with 9 texture primitives are obtained in the segmentation step I, and the texture primitives are segmented from fabric image in the segmentation step II . The segmentation process uses the minimum values on the projection curves. For the rotation invariance of the segmentation, several pixels are added around the segmented sub-regions and texture primitives, and than the segmentation rules are used for further segmentation.

\begin{figure}[H]
\centering
\subfigure[Before smoothing]{
\label{Fig.sub.1}
\includegraphics[scale=.4]{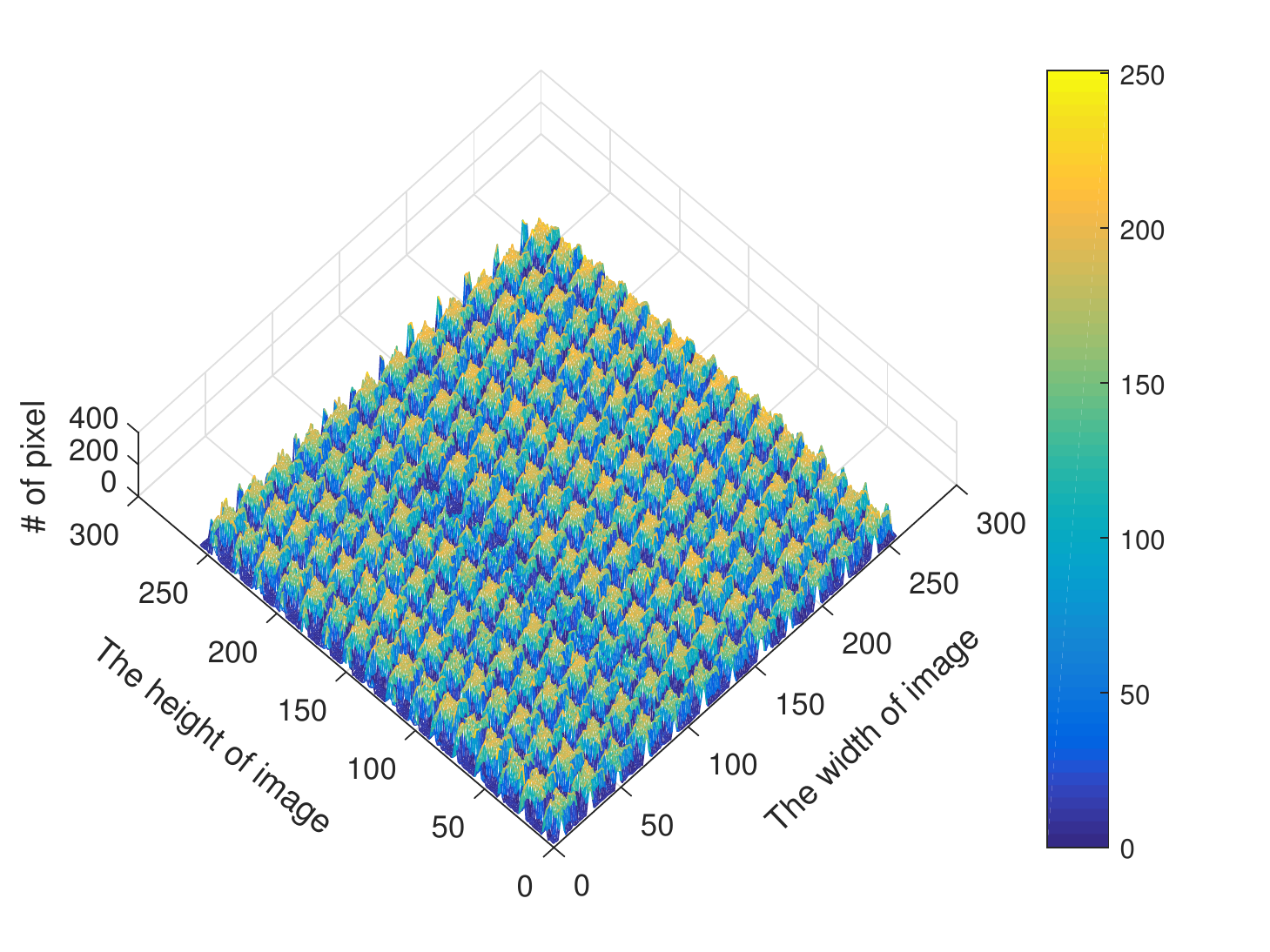}}
\subfigure[After smoothing]{
\label{Fig.sub.2}
\includegraphics[scale=.4]{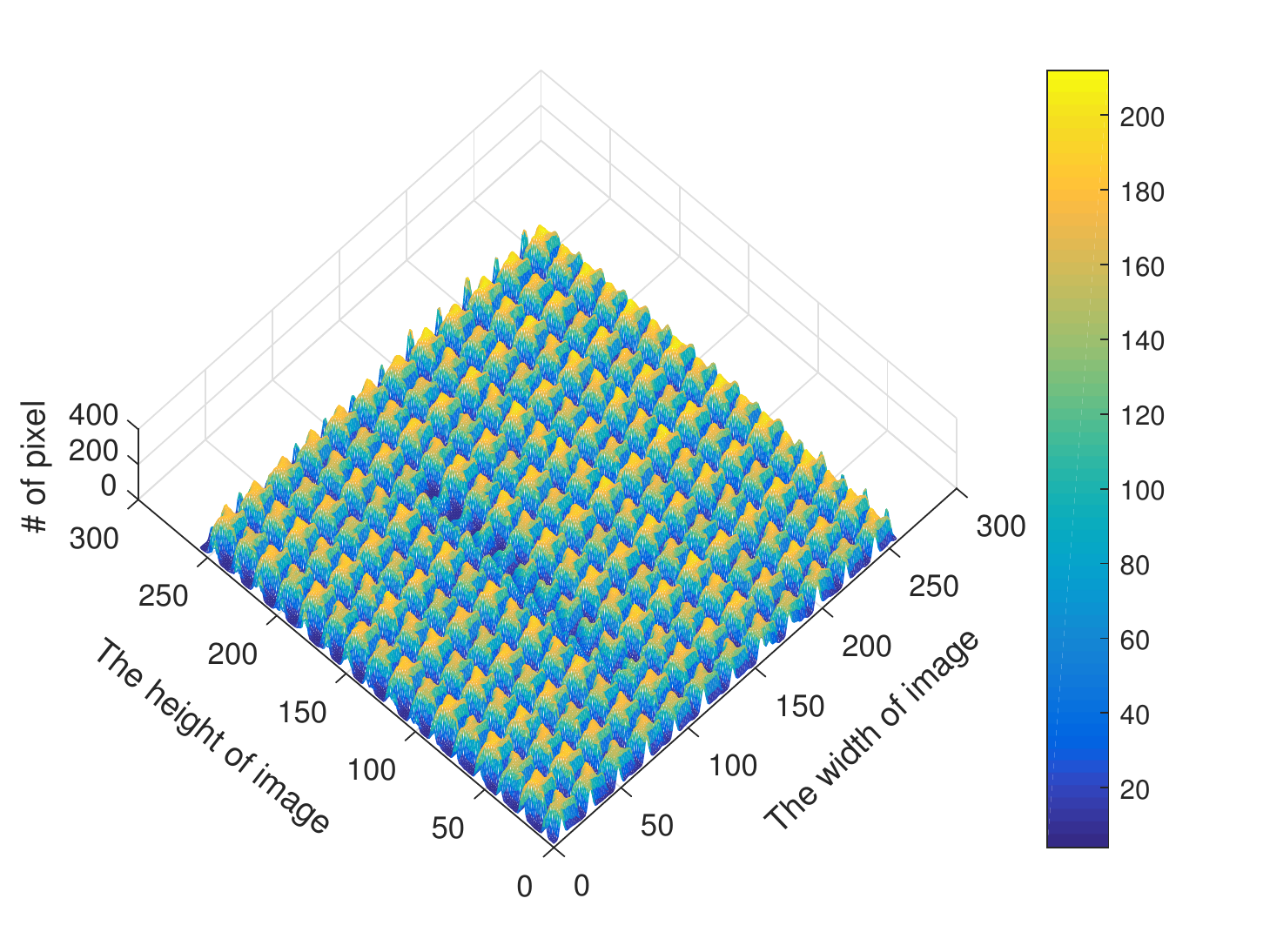}}
\caption{Mesh diagrams of star-patterned fabric.}
\end{figure}

\begin{figure*}[t]
\centering
\includegraphics[scale=.8]{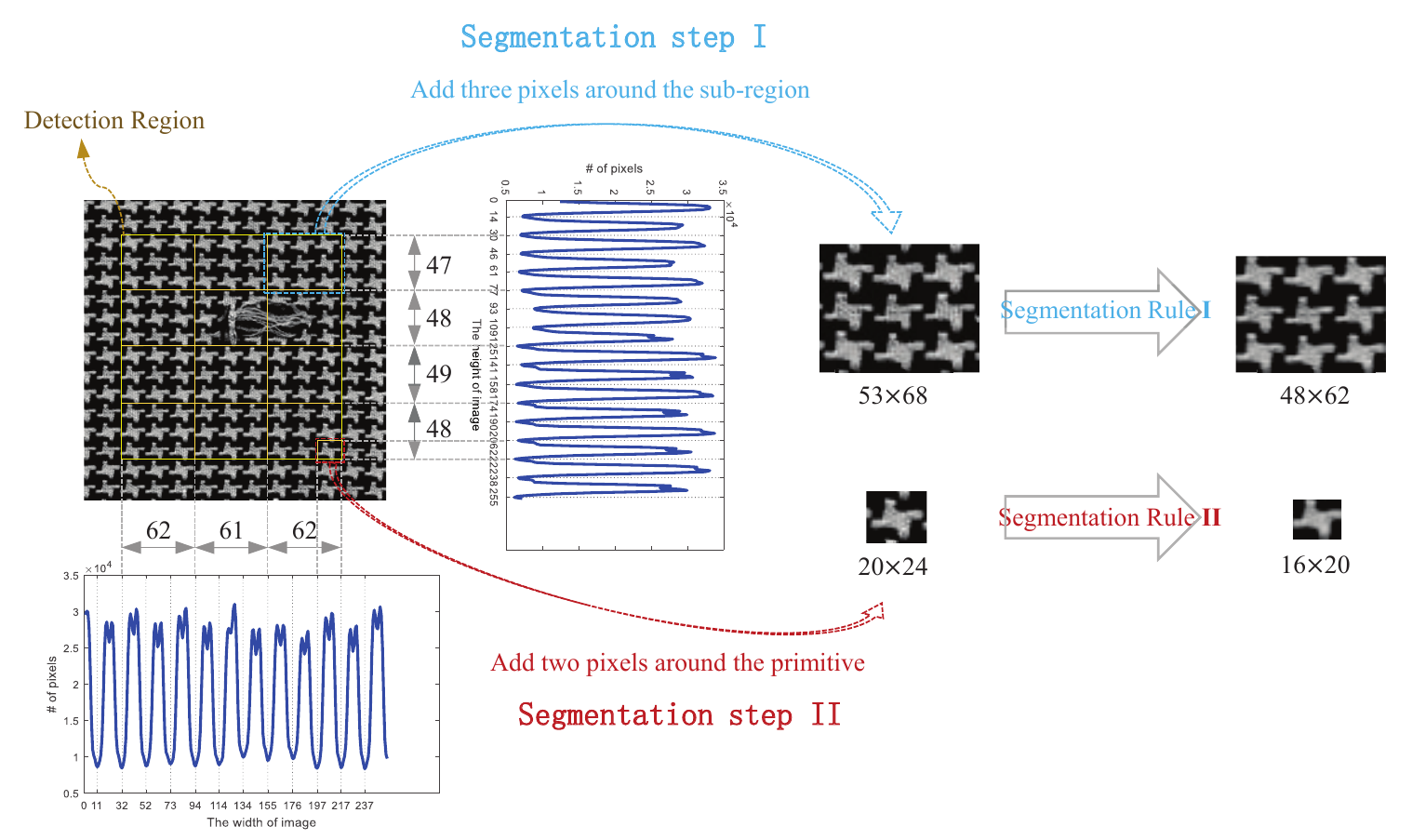}
\caption{Segmentation process of star-patterned fabric.}
\end{figure*}

As shown in Fig. 6, there are two types of template sub-regions and texture primitives. The template sub-regions, $48 \times 62$, are classified by the order of texture primitives. The texture primitives arranged by row in subregion-1 is star-1, star-2 and star-1, and subregion-2 is star-2, star-1 and star-2. Additionally, star-1 and star-2, $16 \times 20$, are distinguished by using the minimum value of vertical projection curve of sub-region. The minimum value of star-1 is bigger than the star-2. Based on the introduction above, the two segmentation rules are described as follow:

(i) As shown in Fig. 5, the segmented subregions with different size before using the segmentation rule I, which can be classified by $l_h, h_1, h_2$ and $l_v, v_1, v_2$. According to our statistics, the range of $l_h$ is 46, 47, 48, 49, and the range of $l_v$ is 35, 36, 37. In addition, different $l_h$ and $l_v$ are with different $h_1, h_2$ and $v_1, v_2$. For example, when $l_h$ is 48, $h_1, h_2$ are 6, 7. When $l_v$ is 37, $v_1, v_2$ are 5, 5. Therefore, each template sub-region has 12 different sizes, and the A-HOG features in each size of sub-region are clustered to obtain the label embedding dictionary. After the segmentation rule I, the further segmented sub-regions are used for AHOG-Net defect detection.

(ii) In the segmentation rule II, the minimum values of projection curves of texture primitive can be easily obtained. According to our statistics, the length around the minimum value of the projection curve is fixed. As shown in Fig. 6, the fixed length in the horizontal and vertical directions is 9, 10 and 7, 8. After the segmentation rule II, the further segmented texture primitives are used for SIFT-Net defect detection.

\begin{figure}[t]
\centering
\includegraphics[scale=.4]{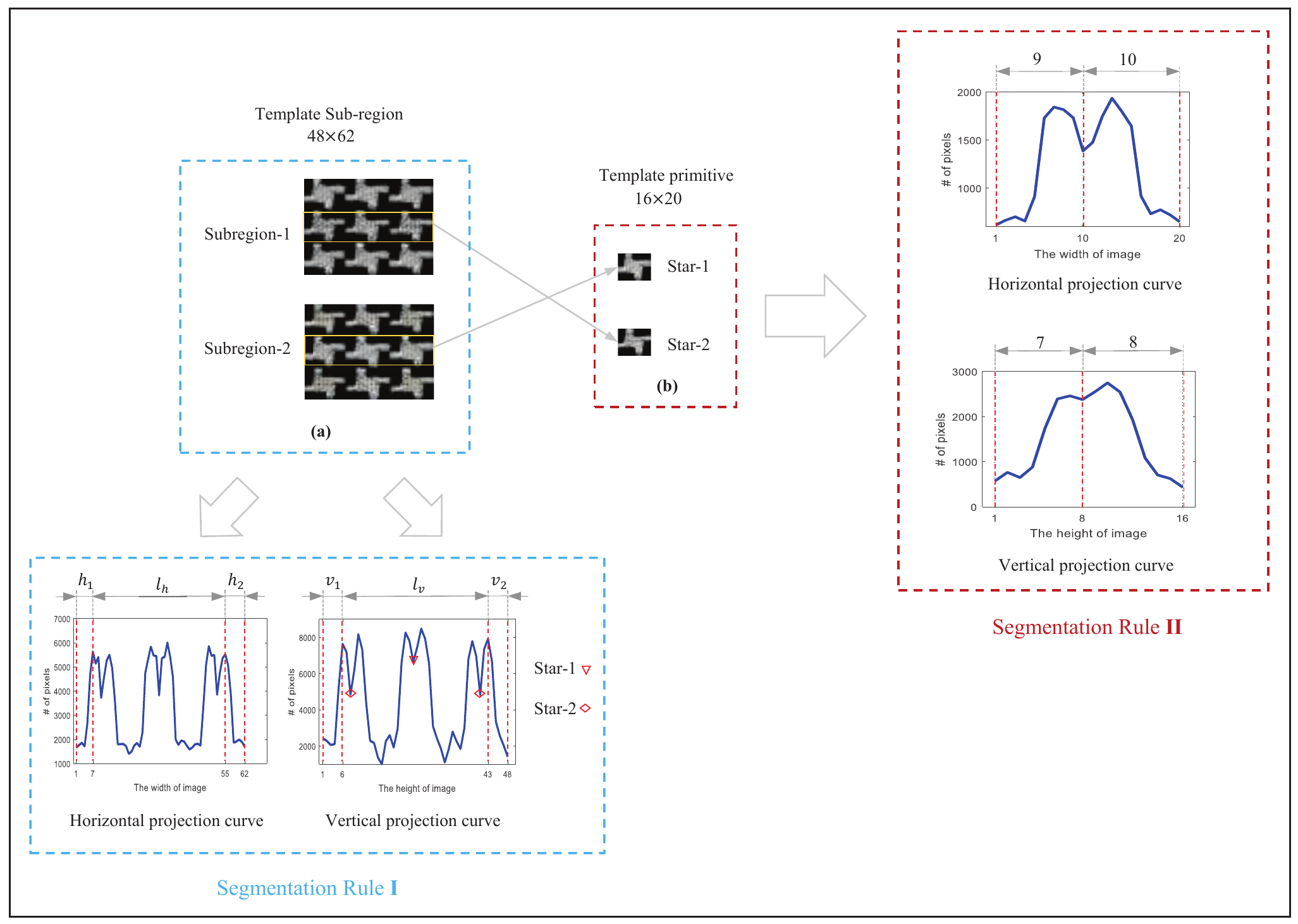}
\caption{Segmentation rule of star-patterned fabric. (a) Two types of template Sub-regions: Subregion-1 and Subregion-2. (b) Two types of template primitives: star-1 and star-2. }
\end{figure}

\subsection{AHOG-Net}
In the AHOG-Net, we use the Aggregated Histograms of Oriented Gradients (A-HOG) feature and the proposed Locality-constrained Reconstruction Error (LCRE) value for fabric defect detection. A-HOG feature is a high dimensional vector, which is the aggregation of local HOG features in image sub-region. As shown in Fig. 1, the A-HOG features is extracted in the sub-regions of fabric image. For enhance the feature representation capability, the Locality-assignment Coding (LSC) is utilized to get the \textit{k}-nearest visual words $\mathcal{N}_k(\textbf{\textit{x}}_i)$ of extracted A-HOG feature and their weights ${\textit{c}}_{ij}$. The visual words are obtained by using the K-Means algorithm to cluster the A-HOG features extracted from normal fabric images. The segmented sub-regions with different $l_h$ and $l_v$ corresponds to different label embedding dictionary $D_z$. $z$ is the number of trained label embedding sub-dictionary. And then the residual vectors, $\exp(-\beta\|\textbf{\textit{x}}_i - \textbf{\textit{b}}_j\|^2)$, between the A-HOG feature and \textit{k}-nearest visual words are calculated. As shown on the left of Fig. 6, we use the \textit{k}-nearest visual words and its weights for image reconstruction. In the LSC encoding, visual word with a larger residual value (closer visual word) gets a larger weight ${\textit{c}}_{ij}$, which accounts for a bigger proportion in the image reconstruction. In this case, the reconstruction coefficient ${\textit{c}}_{ij}$ of closer visual words is larger, which not only conforms to the proportion of different visual words in the image reconstruction, but also preserves the local characteristics of image reconstruction. Therefore, we propose the Locality-constrained Reconstruction Error (LCRE) for fabric defect detection. The LCRE $\varepsilon$ can be calculated as follows:

\begin{equation}
\begin{split}
&{}\varepsilon = \sum_{j=1}^{k} {\textit{c}}_{i,j} \times K_\beta(\textbf{\textit{x}}_i, \textbf{\textit{b}}_j)\\
&{}s.t. \ {\textbf{1}^\top}{{\textit{c}}_{i,j}} = 1, \forall{i}
\end{split}
\end{equation}
where function $K_\beta(\textbf{\textit{x}}_i, \textbf{\textit{b}}_j)$ is calculated from equation (4).

\subsection{SIFT-Net}
As shown in Fig. 1, the SIFT-Net includes training state and testing state. (i) The training state is aim to obtain the number embedding dictionary and template binary array. Firstly, dense SIFT descriptors \cite{Lowe2004} are extracted from each patch of normal texture primitives in a certain stride. Then, the SIFT descriptors in the same patch is clustered by K-Means to obtain a visual word. Finally, the visual words are randomly numbered to form the number embedding dictionary. (ii) In the testing state, the feature extraction layer is same as the training state. The feature representation layer includes our proposed Restrictive Locality-constrained Coding (RLC) and Bag-of-Indexes (BoI) methods. In the following subsection, the proposed methods and different layers of testing state in the SIFT-Net are introduced.

\subsubsection{Restrictive Locality-constrained Coding}
The proposed Restrictive Locality-constrained Coding (RLC) consider the distance between the descriptor and its k-nearest visual words, which can achieve much less quantization error than LLC. As shown on the right of Fig. 6, only the visual words within the local scope are valid for encoding in RLC. Otherwise, encoding is invalid. We use the RLC approach in the feature representation layer to extract more distinctive feature for fabric defect detection. The proposed RLC method encodes each local descriptor $\textbf{\textit{x}}_i$ by solving the following problem:
\begin{equation}
\begin{split}
&{}\mathop{\min}_{\tilde{C}} \sum\limits_{i=1}^{N} \ \|\ {\textbf{\textit{x}}_i - {\hat{\textbf{\textit{B}}}_i}\tilde{\textbf{\textit{c}}}_i\ \|}^2 \\
&{}s.t. \ \textit{dist}(\textbf{\textit{x}}_i, \hat{\textbf{\textit{B}}}_i) < \rho \\
&{}\qquad {\textbf{1}^\top}{\tilde{\textbf{\textit{c}}}_i} = 1, \forall{i} \\
\end{split}
\end{equation}
\begin{equation}
\rho = \mathop{\min}_{1 \leq i \leq N} \textit{dist}(\textbf{\textit{x}}_i, \hat{\textbf{\textit{B}}}_i)
\end{equation}
where dictionary $\hat{\textbf{\textit{B}}}$ is obtained during the training state shown on the left of Fig. 1. $\rho$ is the minimum distance between local descriptors of normal fabric images and its corresponding visual words, which indicates the effective scope of locally encoding shown on the right of Fig. 6.

\begin{figure}[t]
\centering
\includegraphics[scale=.6]{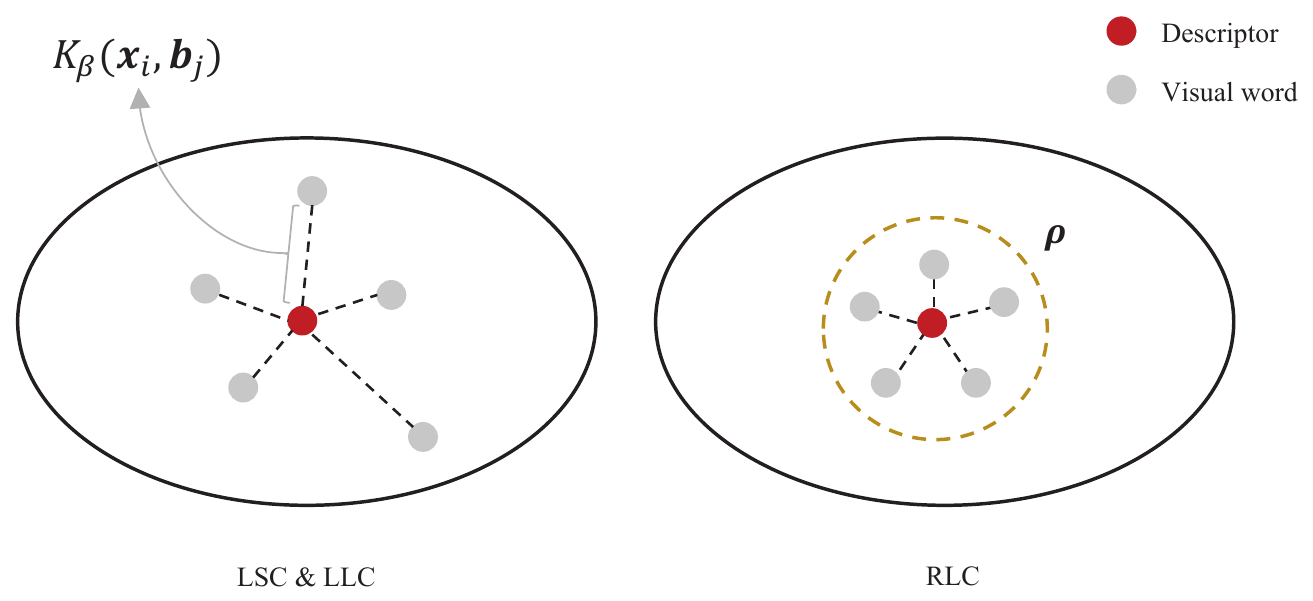}
\caption{Comparison between LSC, LLC and RLC.}
\end{figure}

\subsubsection{Bag-of-Indexes}
RLC coding and feature map based on proposed Bag-of-Indexes (BoI) method together form the feature representation layer. BoI method uses the indexes of visual words for feature map. The differences between our proposed BoI method and the BoF method is:
\begin{itemize}
\item In the training state, visual words generation is repeated in patch of the normal texture primitives, instead of clustering all local descriptors at once. Additionally, the visual words are randomly numbered in BoI method.
\item BoI method uses the order of visual words to overcome the disadvantage of unordered appearance descriptors in
      BoF method. In section 4, our experiments show that BoI method performs better than the BoF method.
\end{itemize}

\subsubsection{The full connection layer}
Through the feature representation layer with RLC coding and feature map based on BoI method, the BoI feature is obtained. In the full connection layer, the BoI feature is fully connected to form the BoI vector. One-hot encoding is widely used in Natural Language Processing (NLP) \cite{Rong2014}, it can vectorize discrete words for semantic analysis. In our method, each BoI feature has \textit{M} possible values, \textit{M} is the size of dictionary, and after one-hot encoding, it becomes \textit{M} sparse binary features. And only one visual word is activated at a time, corresponding to "1" in the binary array. There is no connection between "1", which means that the dependencies between local descriptors are not considered. If an image is treated as a sentence, the different combinations of local descriptors (words) represent different objects (semantics).

To some extent, different coding methods, such as soft-assignment, LSC, and LLC etc., consider the connection between local descriptors. However, the subsequent use of max-pooling only preserves the maximum response of local descriptors on the visual words, which causes the association between the local descriptors to be lost. As our experiments in section 4 conclude that the index of visual words plays an major role in the coding methods.

\subsubsection{The defect detection layer}
In the SIFT-Net, defect detection layer includes two steps. Let the template binary table and the binary array of testing fabric image be $\textbf{\textit{T}}_1$, $\textbf{\textit{T}}_2$. In the first detection step, the Hamming distance, $\textit{Dif}$, between $\textbf{\textit{T}}_1$, $\textbf{\textit{T}}_2$ is calculated:
\begin{equation}
\textit{Dif} \ (\textbf{\textit{T}}_1, \textbf{\textit{T}}_2) = \sum_{i=1}^{N} \sum_{j=1}^{N} \textbf{\textit{T}}_1(i,j) \oplus \textbf{\textit{T}}_2(i,j)
\end{equation}
where $\oplus$ indicates "XOR" operator. If $\textit{Dif}$ isn't zero, the test fabric is defective. Otherwise, the second detection step is used. The distance between descriptor and its nearest visual word, $\textit{dist}(\textbf{\textit{x}}_i, \hat{\textbf{\textit{B}}}_i)$, compares with the threshold $\rho$. If $\textit{dist}(\textbf{\textit{x}}_i, \hat{\textbf{\textit{B}}}_i)$ is less than the $\rho$, the test fabric is defect-free. Otherwise, the fabric is defective.

\section{Experimental results}
\subsection{Dataset}
To evaluate the performance of proposed CZI-Net, two sets of patterned fabrics (star-patterned and box-patterned) are used for defect detection. The Fabric Image Dataset (FID) is provided by Industrial Automation Research Laboratory from Dept. of Electrical and Electronic of Hong Kong University. There are totally 50 including 25 normal and 25 defective fabric images in each patterned class. The types of defective fabrics are BrokenEnd, Hole, NettingMultiple, ThickBar, and ThinBar. The example patterned fabric images in each type are illustrated in Fig. 8.

After the segmentation process, the star-patterned fabrics get 600 sub-regions (12 sub-regions per fabric image) for AHOG-Net detection and 5400 (9 primitives per sub-region) texture primitives for SIFT-Net detection. All the experiments are implemented on an Intel Core i7-8700K 3.7GHz CPU with 16G memory.

\begin{figure}[t]
\centering
\includegraphics[scale=.58]{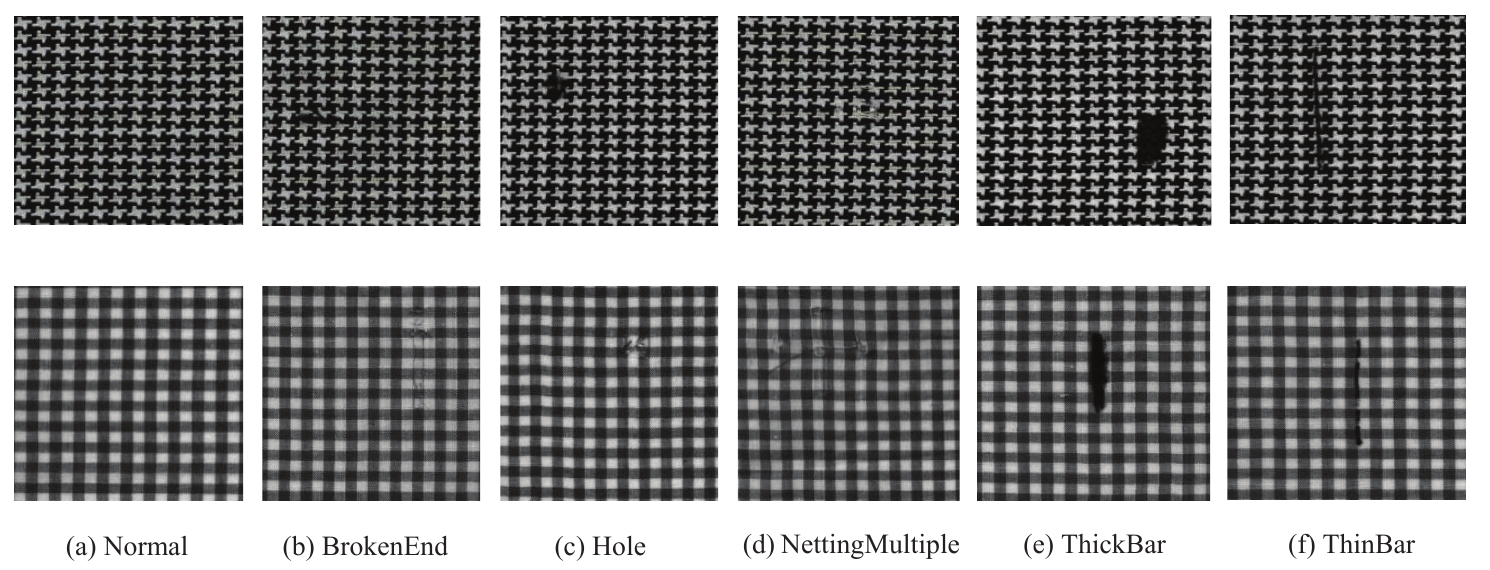}
\caption{The example fabric images of box-patterned and star-patterned.}
\end{figure}

\subsection{AHOG-Net}
In the feature extraction layer, the block of HOG features extraction is $8 \times 10$, and the cell is $4 \times 5$. The gradient direction of the cell is divided into 9 directions. Therefore, the dimension each HOG feature is 36, and the dimension of A-HOG feature is 1296. In the feature representation layer, LSC coding method \cite{LiuWangLiu2011} is used to get the \textit{k}-nearest visual words of A-HOG features for image construction. In our experiment, $k$ is 3, and the label embedding dictionary is 5. In the defect detection layer, the AHOG-Net includes training state and testing state for fabric defect detection. In the training state, the threshold value of LCRE is obtained from 300 normal sub-regions of normal fabric. From Fig. 9, we can observe that the threshold of RCLE can be set as 5.

\begin{figure}[t]
\centering
\includegraphics[scale=.52]{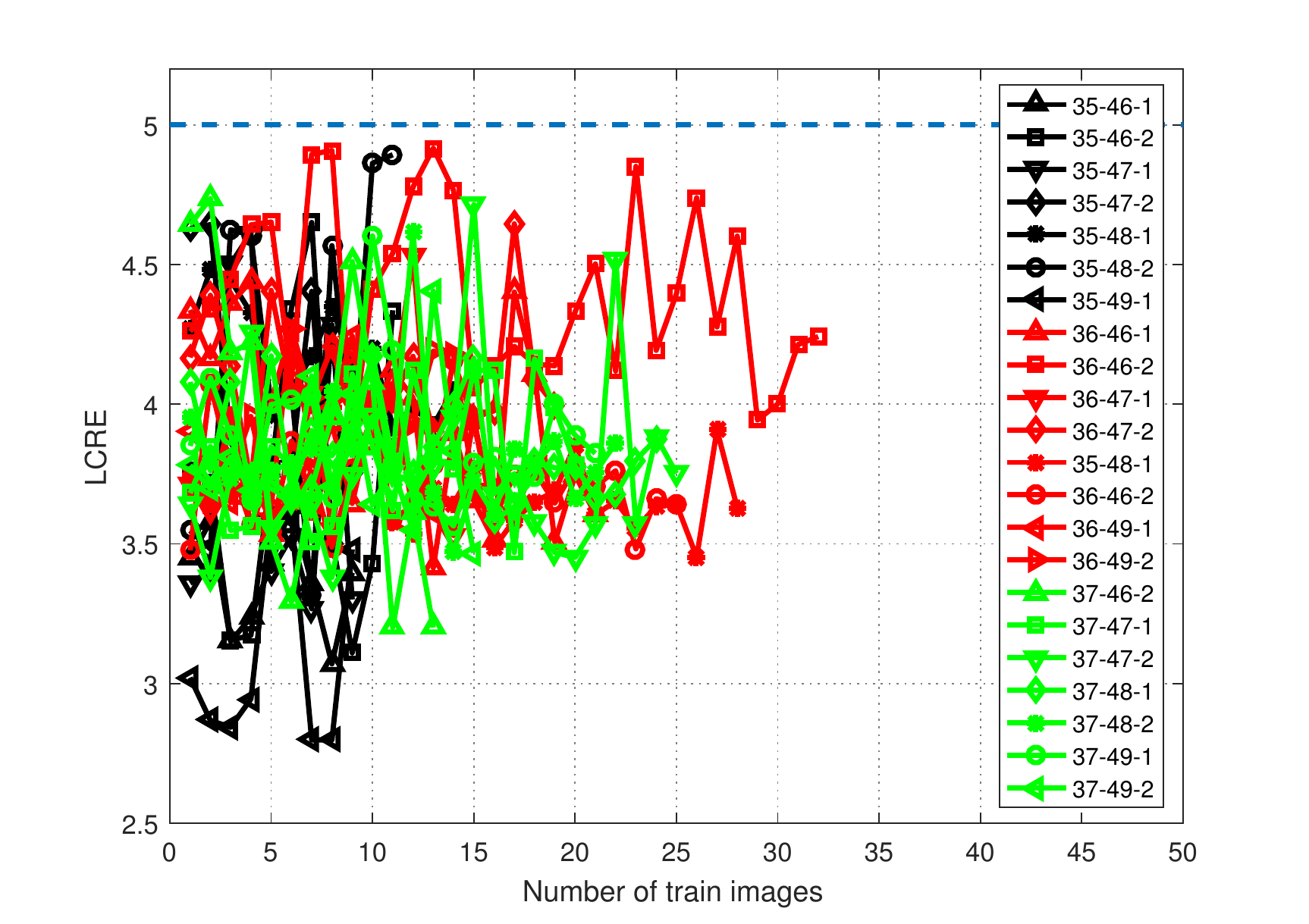}
\caption{The threshold value of LCRE. In the figure, 36-48 means that $l_h$ and $l_v$ are 36 and 48, and "-1" and "-2" indicates the types of sub-region: subregion-1 and subregion-2.}
\end{figure}

In the testing state, the experimental results of the other 300 sub-regions with 235 normal and 65 defective fabrics are shown in Fig. 10. We can conclude that the detection accuracy of AHOG-Net is 98.33\%, recall rate is 96.875\%.

\begin{figure}[t]
\centering
\subfigure[]{
\label{Fig.sub.1}
\includegraphics[scale=.4]{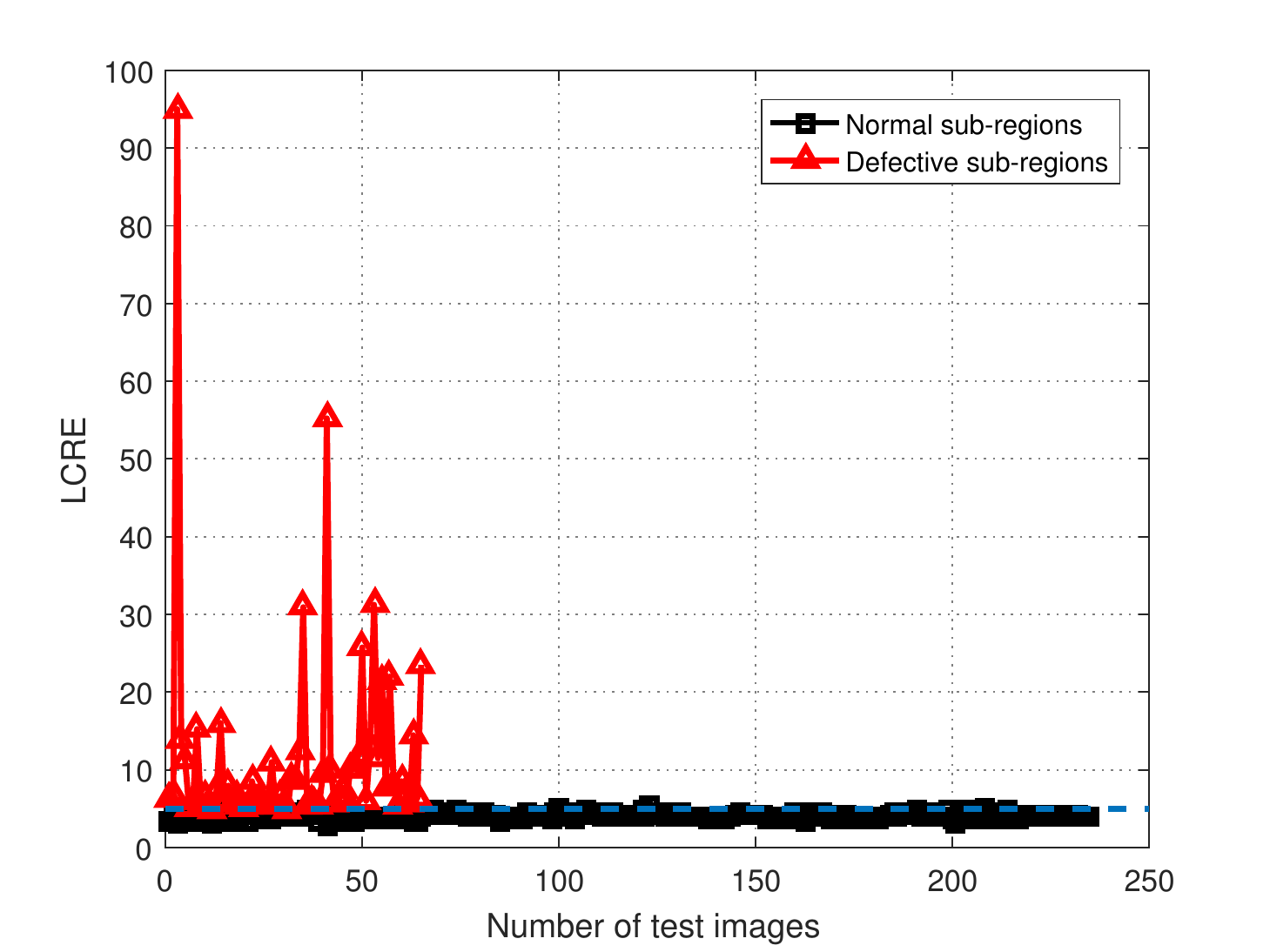}}
\subfigure[]{
\label{Fig.sub.2}
\includegraphics[scale=.4]{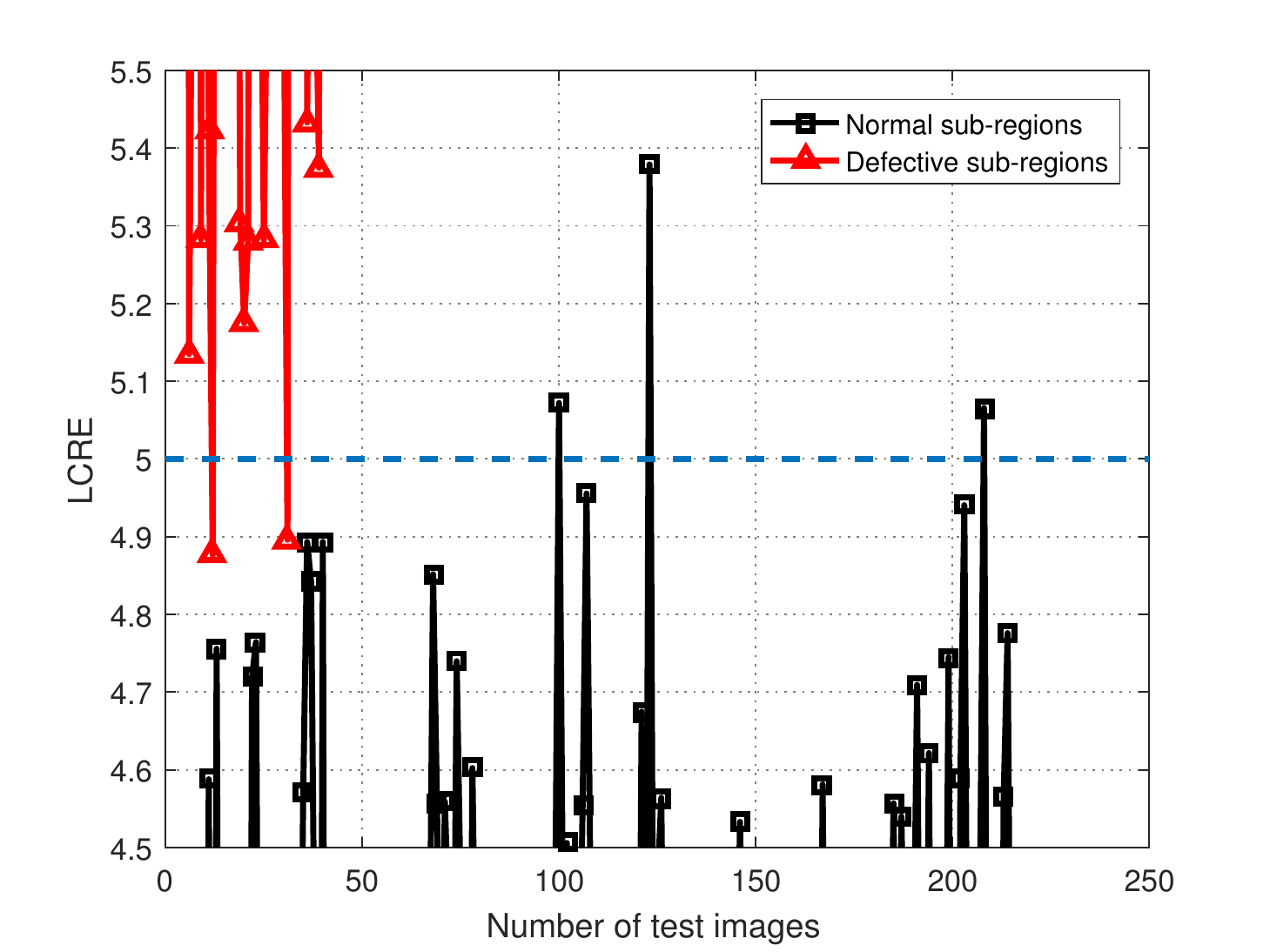}}
\caption{Defect detection performance of AHOG-Net. (b) is the drawing of partial enlargement of (a).}
\end{figure}

\subsection{SIFT-Net}
In the feature extraction layer, the patch of SIFT descriptors extraction is $16 \times 16$, and the patches densely located by every 2 pixels on the fabric image. Therefore, 3 SIFT descriptors are extracted from a texture primitive, and its dimension is 128. In the defect detection layer, the SIFT-Net also includes training state and testing state. In the training state, 100 normal primitives from normal fabrics are used to train the number embedding dictionary, and then the template binary array and the threshold value of locally encoding scope are obtained based on the trained dictionary. In our experiments, we conclude that the template binary array $\textbf{\textit{T}}_1$ is:
$$\textbf{\textit{T}}_1 =
  \left[ \begin{array}{ccc}
   1 & 0 & 0  \\
   0 & 0 & 1  \\
   0 & 1 & 0
  \end{array}
  \right ]$$
The minimum of $\textit{dist}(\textbf{\textit{x}}_i, \hat{\textbf{\textit{B}}}_i)$ is shown in Fig. 11. From Fig. 11, we can observe that the threshold of Hamming distance is 0, and the threshold of maxmum distance between SIFT descriptors and its nearest visual word is 0.12.

\begin{figure}[t]
\centering
\subfigure[]{
\label{Fig.sub.1}
\includegraphics[scale=.4]{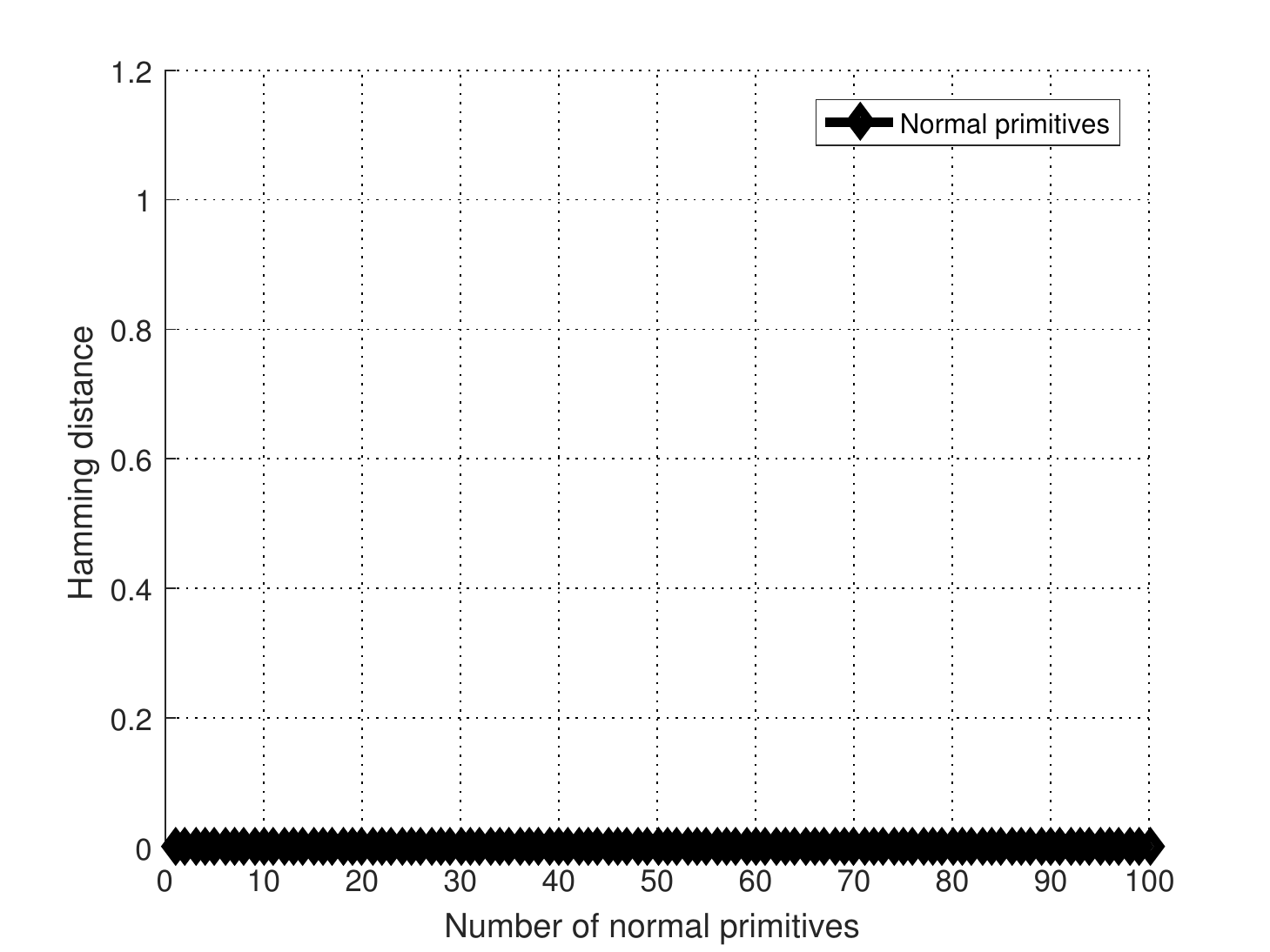}}
\subfigure[]{
\label{Fig.sub.2}
\includegraphics[scale=.4]{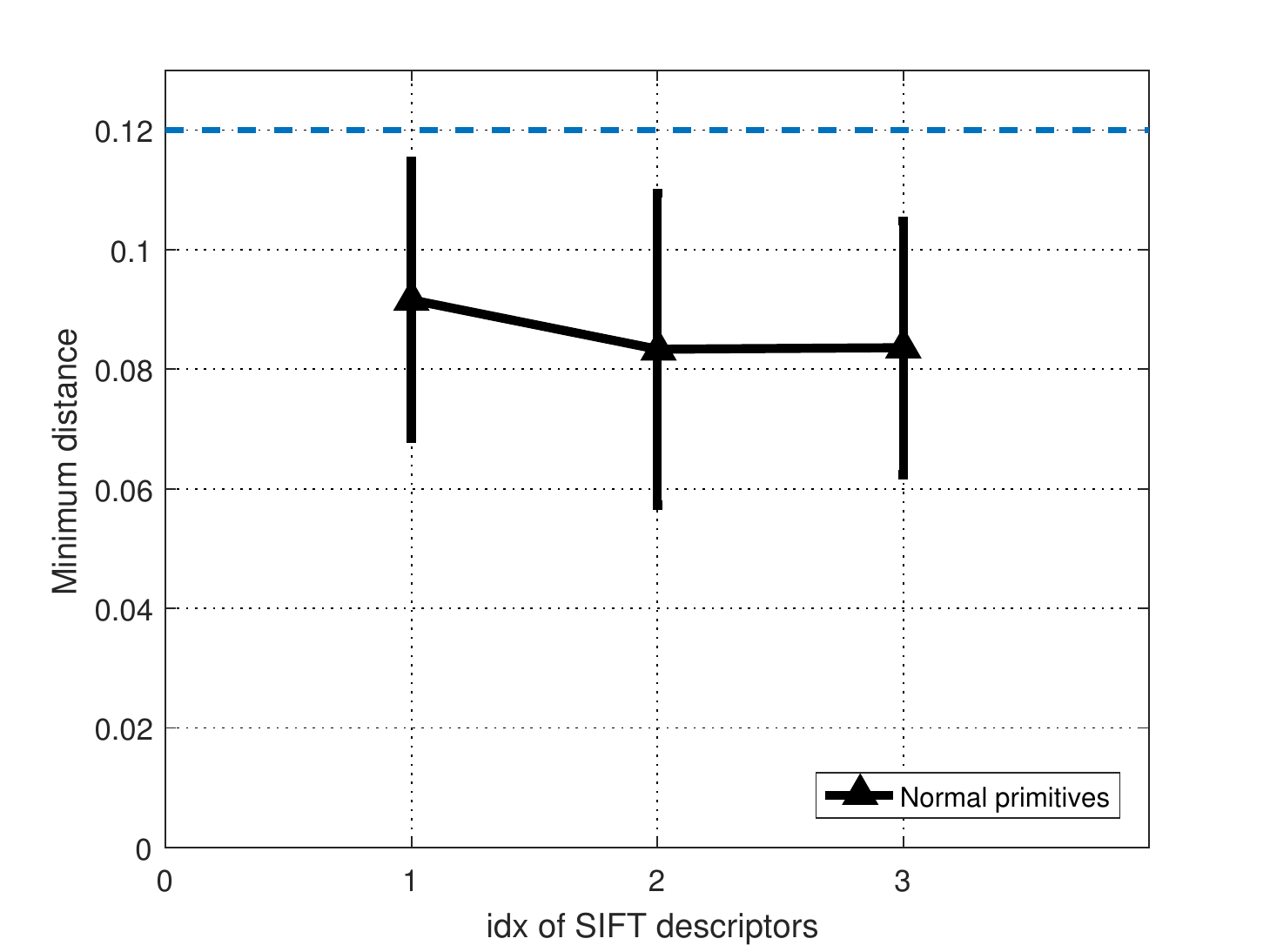}}
\caption{Training state of SIFT-Net. (a) The Hamming distance between template binary array and binary array of trained primitives. (b) The maximum distance between SIFT descriptors and its nearest visual word.}
\end{figure}

In the testing state, texture primitives in the defective sub-regions will be detected for location determination in the SIFT-Net. From the observation of Fig. 10, there are 66 sub-regions (594 texture primitives) that AHOG-Net has determined that they are defective. According to our statistics, there are 380 normal primitives and 214 defective primitives for SIFT-Net defect detection. The defect detection layer of SIFT-Net includes two step, and the experimental results of the first step is shown in Fig. 12. From Fig. 12 (a), we can observe that the Hamming distance of 42 defective primitives aren't zero, which means they are correctly detected. The remainder primitives will be detected in the second step of detection layer. As shown in the (b) and (c) of Fig. 12, we can observe that the distance between SIFT descriptors of normal primitives and its nearest visual word is less then the $\rho$ threshold 0.12, which means that all normal primitives are correctly detected in the second step of SIFT-Net. Therefore, the false alarm rate 3.125\% of AHOG-Net can be eliminated. However, the location of 17 defective primitives aren't correctly detected, the detection accuracy of the second detection step is 97.14\%.

\begin{figure}[t]
\centering
\subfigure[First detection step]{
\label{Fig.sub.1}
\includegraphics[scale=.5]{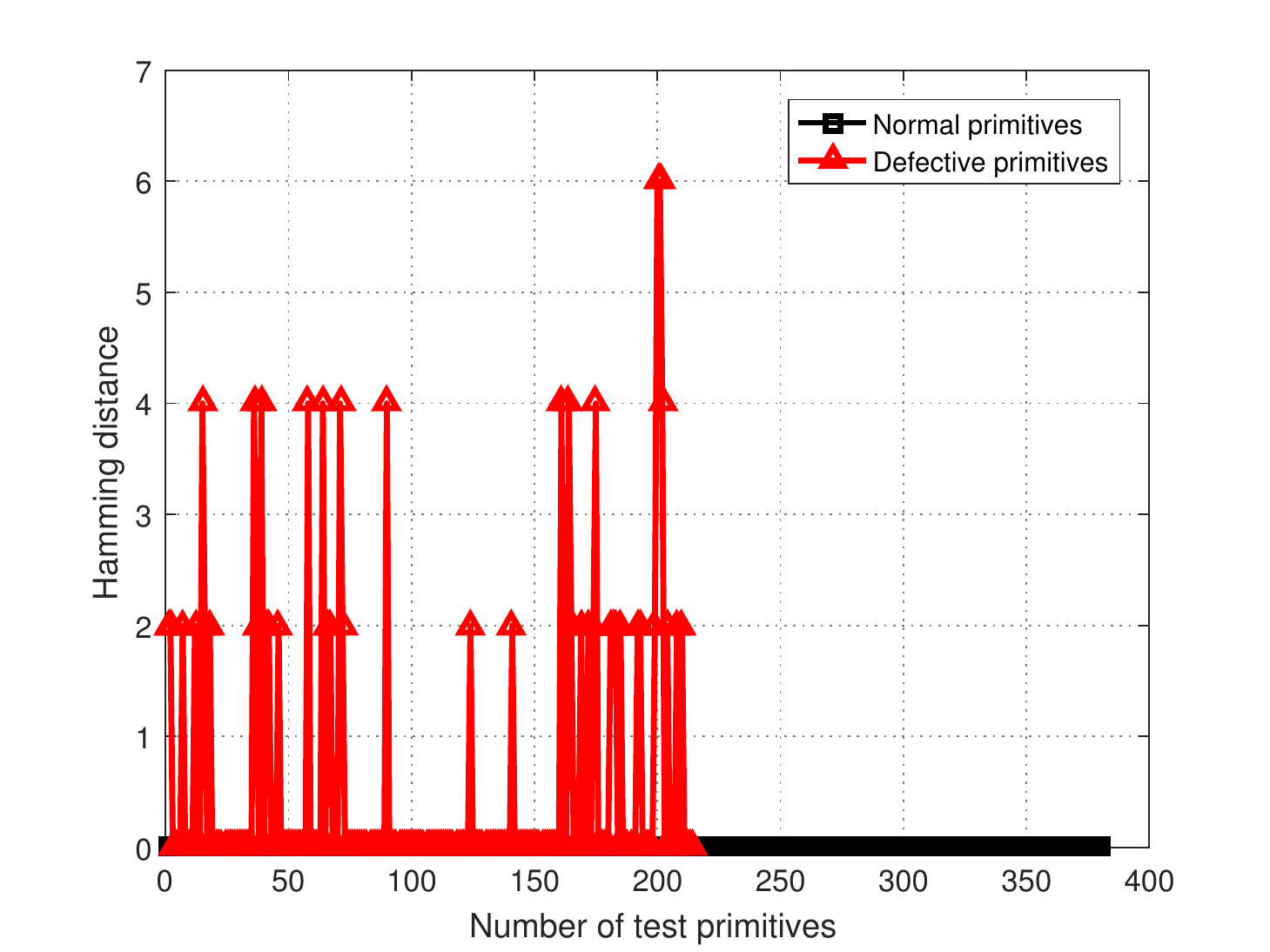}}
\subfigure[Second detection step]{
\label{Fig.sub.1}
\includegraphics[scale=.25]{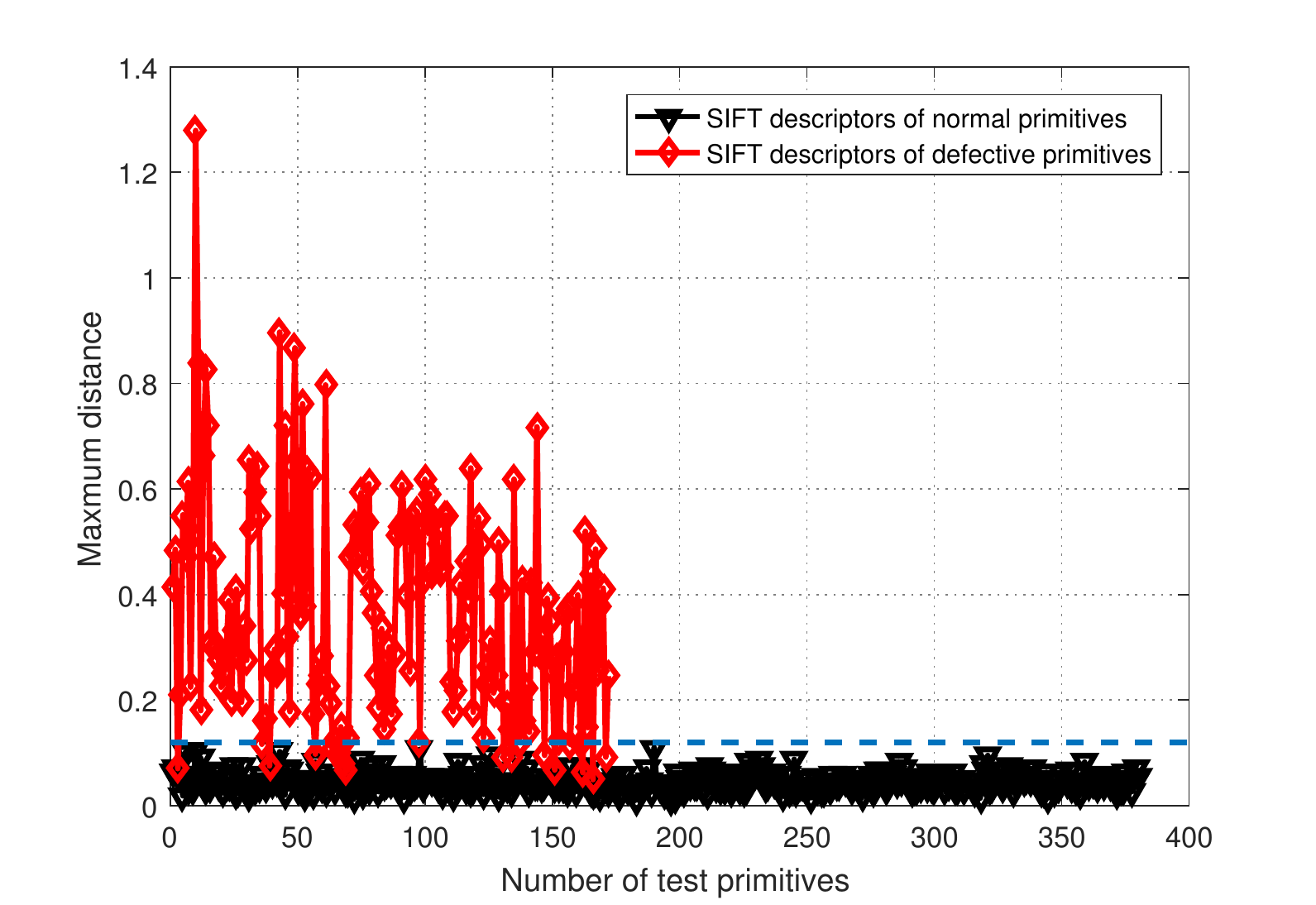}}
\subfigure[]{
\label{Fig.sub.2}
\includegraphics[scale=.25]{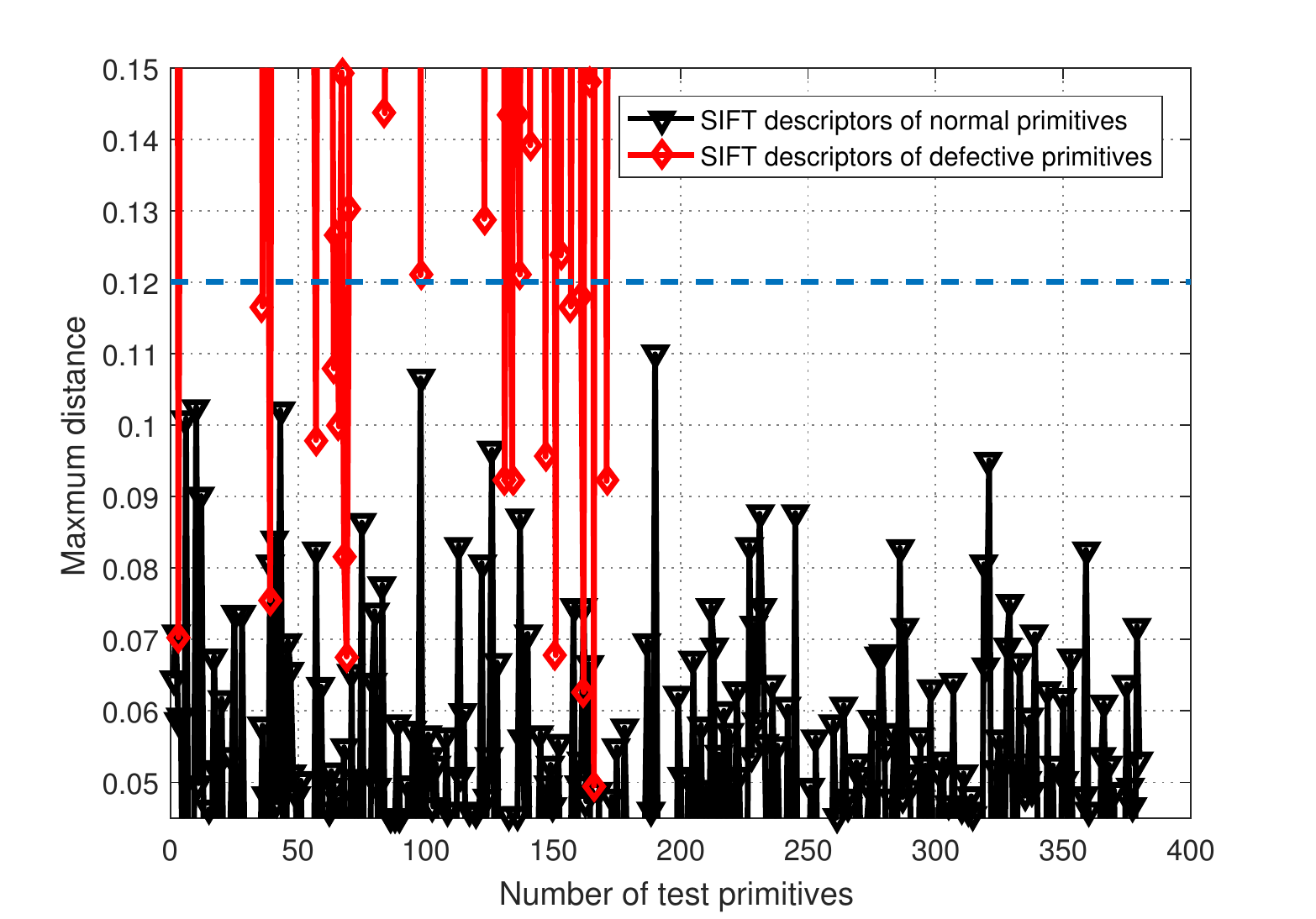}}
\caption{Testing state of SIFT-Net. (a) The Hamming distance between template binary array and binary array of tested primitives. (b) The maximum distance between SIFT descriptors and its nearest visual word. (C) is the drawing of partial enlargement of (b).}
\end{figure}

To provide more convincing detection performance of our CZI-Net, Fig. 13 shows the receiver operating characteristic (ROC) curve for 10 replicate experiments. Based on the above analysis, the detection accuracy of sub-regions is 99.34\%. From Fig. 10, we can observe that the two not detected sub-regions are not in the same fabric (they are not close together in Fig. 10), so the detection accuracy of proposed CZI-Net is 100\% for each detected fabric image. Table I shows the star-patterned fabric defect detection results of different methods. Compared with the DCNNs methods, our CZI-Net achieves higher accuracy with less detection time. Compared with the method of Y. t. Ngan (2009), our method achieves considerable detection accuracy, and the advantages of our method is that the locations of defects can be accurately determined. Moreover, the detection time of proposed CZI-Net is about 180ms per fabric image, which includes the image segmentation, AHOG-Net detection and SIFT-Net detection. Compared with DCCNs based methods, such as \cite{LiZhaoPan2017}, our method doesn't require a lot of data to train the network. The final detection results on star-patterned fabric are shown in Fig. 14.

\begin{figure}[t]
\centering
\includegraphics[scale=.55]{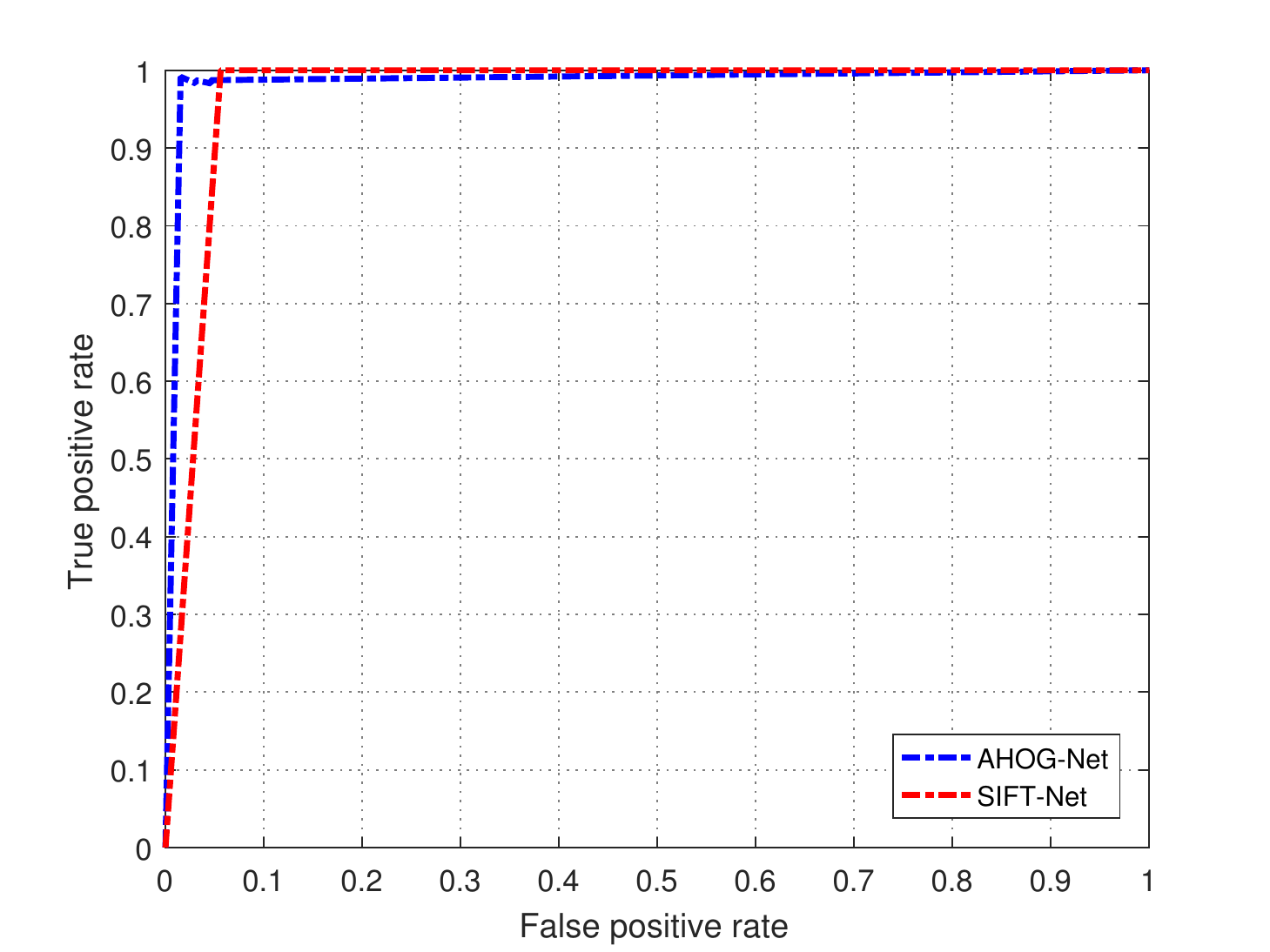}
\caption{The ROC curves for the proposed AHOG-Net and SIFT-Net on star-patterned fabric.}
\end{figure}

\begin{table}[t]
\caption{Star-patterned fabric defect detection results}
\vspace{-0.5cm}
\begin{center}
\begin{tabular}{ccc}
\hline
\textbf{Method} & \textbf{Accuracy}   \\
\hline
\cite{Ngan2009} & \textbf{100}  \\
\cite{Asha2012} & 99.3       \\
\cite{NganYuanZhang2014} & 97.55      \\
\cite{Tsang2016} & 99.07      \\
\cite{Martinez-Leon2016} & 96.8       \\
\cite{MK2017} & 99.07      \\
\cite{LiZhaoPan2017} & 95.83 \\
\cite{Jia2018} & 94         \\
\cite{Hamdi2018} & 92.40      \\
\hline
Ours & \textbf{100}  \\
\hline
\end{tabular}
\end{center}
\end{table}

\begin{figure}[t]
\centering
\includegraphics[scale=.5]{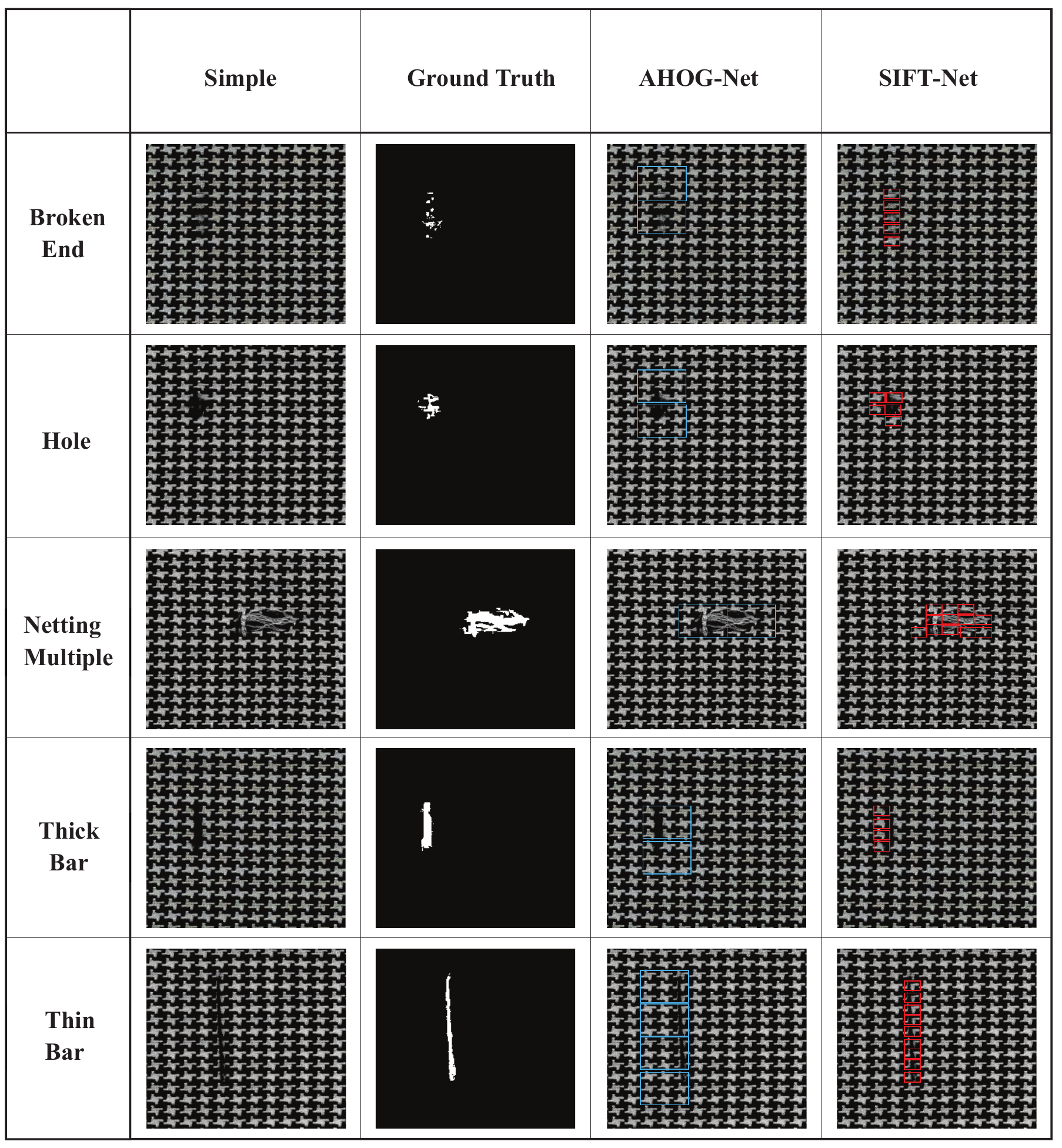}
\caption{Star-patterned fabric defect detection results with our CZI-Net.}
\end{figure}

\subsection{Discussion}
To provide more convincing analysis that the proposed BoI method performs better than the BoF method, we further compared the two method on Caltech-101 image classification dataset \cite{Fei-FeiFergus2007}. The Caltech-101 dataset contains 9144 images in 102 classes including airplanes, faces, watches, etc. The number of images per category varies from 31 to 800. We partitioned the whole dataset into 30 training images per class and the remainder images for testing. In our setup, SIFT descriptors were extracted from patches densely located by every 6 pixels on the image, and the patch is $16 \times 16$. The sub-regions for SPM are $4 \times 4$, $2 \times 2$, $1 \times 1$, and linear SVM is used for image classification.

The number embedding dictionary is obtained by clustering each category image using K-Means. In this way, the clustered visual words in the different category is distributed in the different position of dictionary, and the importance of the indexes of visual words in image classification is more obvious. The dictionary used in our experiments are 510, 1020, 2040 and 4080. However, the fully connected binary array makes curse of dimensionality for image classification. To reduce the dimension, we use the max-pooling following the columns of binary array. The obtained vector is also binary value, "1" and "0" mean the presence or absence of each visual word. Finally, the binary vector is put into linear SVM for image classification. We compared our result with BoF and LLC methods, and the detailed results are shown in Fig. 15.

\begin{figure}[t]
\centering
\includegraphics[scale=.65]{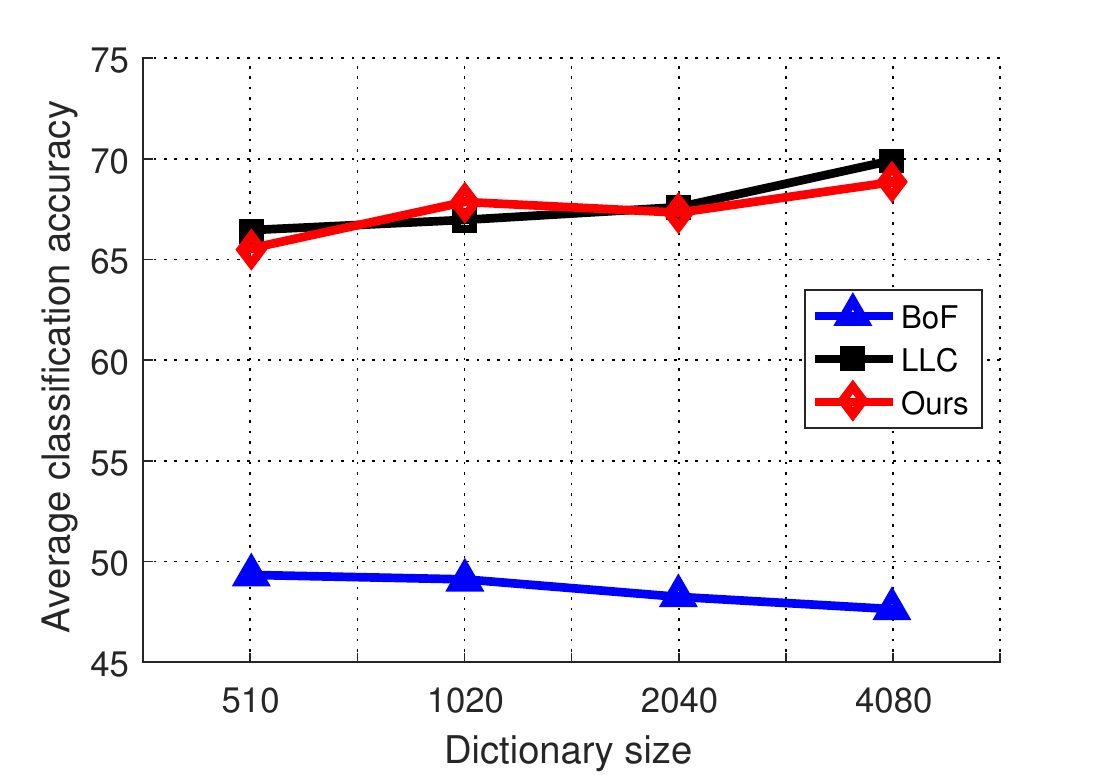}
\caption{Performance of different methods.}
\end{figure}

As shown in Fig. 15, The performance of BoI method is better than the BoF method, and comparable with the LLC method. We can also conclude that the indexes of visual words play a more important in the coding methods. In the LLC method, the max-pooling operations only preserve the maximum response of visual words on the image, which makes the mutual connection of local descriptors lost.

\section{Conclusion}
This paper proposes a two-step Cascaded Zoom-In network (CZI-Net) for fabric defect detection. The AHOG-Net is a computationally simple step while the SIFT-Net can determine the location of defects with high accuracy. Based on CZI-Net, we propose LCRE method in the defect detection layer of AHOG-Net, and RLC, BoI methods in the feature representation layer of SIFT-Net for enhancing the image representation capability. For most defect-free fabrics, only the AHOG-Net was involved. And only those fabrics with high LCRE value was involved in SIFT-Net. Experimental results based on widely used fabric dataset validate state-of-the-art performance of our network. In addition, this paper also presents that the BoI method performs better than the BoF method in image classification, and experiments shows that the index of visual word is more important in the coding methods.


{\small
\bibliographystyle{ieee_fullname}
\bibliography{egbib}
}

\end{document}